%% file: MorphoCluster-Efficient-Annotation-of-Plankton-images-by-Clustering.tex
\documentclass[preprints,article,accept,moreauthors,pdftex]{Definitions/mdpi}

\usepackage{longtable}
\usepackage{siunitx}
\usepackage{csquotes}
\usepackage{cleveref}
\usepackage{caption}
\usepackage{subcaption}
\usepackage{siunitx}
\usepackage{placeins}
\DeclareSIUnit{\nounit}{\relax}
\DeclareMathOperator*{\mean}{mean}

\makeatletter
\g@addto@macro\@floatboxreset\centering
\makeatother

\crefformat{footnote}{#2\footnotemark[#1]#3}

\input{constants.tex}

\firstpage{1} 
\pubvolume{xx}
\issuenum{1}
\articlenumber{5}
\pubyear{2020}
\copyrightyear{2020}
\history{Received: date; Accepted: date; Published: date}

\Title{MorphoCluster: Efficient Annotation of Plankton images by Clustering}

\Author{Simon-Martin Schröder $^{1}$\orcidA{}, Rainer Kiko $^{2,3}$\orcidB{} and Reinhard Koch
$^{1}$\orcidC{}}

\AuthorNames{Simon-Martin Schröder, Rainer Kiko and Reinhard Koch}

\address{%
$^{1}$ \quad Kiel University\\
$^{2}$ \quad Laboratoire d'Océanographie de Villefranche-sur-mer\\
$^{3}$ \quad GEOMAR Helmholtz Center for Ocean Research Kiel%
}

\corres{Correspondence: sms@informatik.uni-kiel.de}

\abstract{In this work, we present MorphoCluster, a software tool for data-driven, fast and accurate annotation of large image data sets.
While already having surpassed the annotation rate of human experts, volume and complexity of marine data will continue to increase in the coming years. Still, this data requires interpretation.
MorphoCluster augments the human ability to discover patterns and perform object classification in large amounts of data by embedding unsupervised clustering in an interactive process.
By aggregating similar images into clusters, our novel approach to image annotation increases consistency, multiplies the throughput of an annotator and allows experts to adapt the granularity of their sorting scheme to the structure in the data.
By sorting a set of 1.2M objects into 280 data-driven classes in 71 hours (16k objects per hour), with \SI{90}{\percent} of these classes having a precision of \num{0.88888888} or higher.
This shows that MorphoCluster is at the same time fast, accurate and consistent, provides a fine-grained and data-driven classification and enables novelty detection.
MorphoCluster is available as open-source software at \url{https://github.com/morphocluster}.}


\begin{document}

\input{Introduction.tex}

\section{Methods}%
\label{sec:methods}

In this section, we present the overall structure of the MorphoCluster approach and the details of our implementation.

\subsection{General overview of the MorphoCluster process}

\begin{figure}[tp]
	\includegraphics[width=0.5\textwidth]{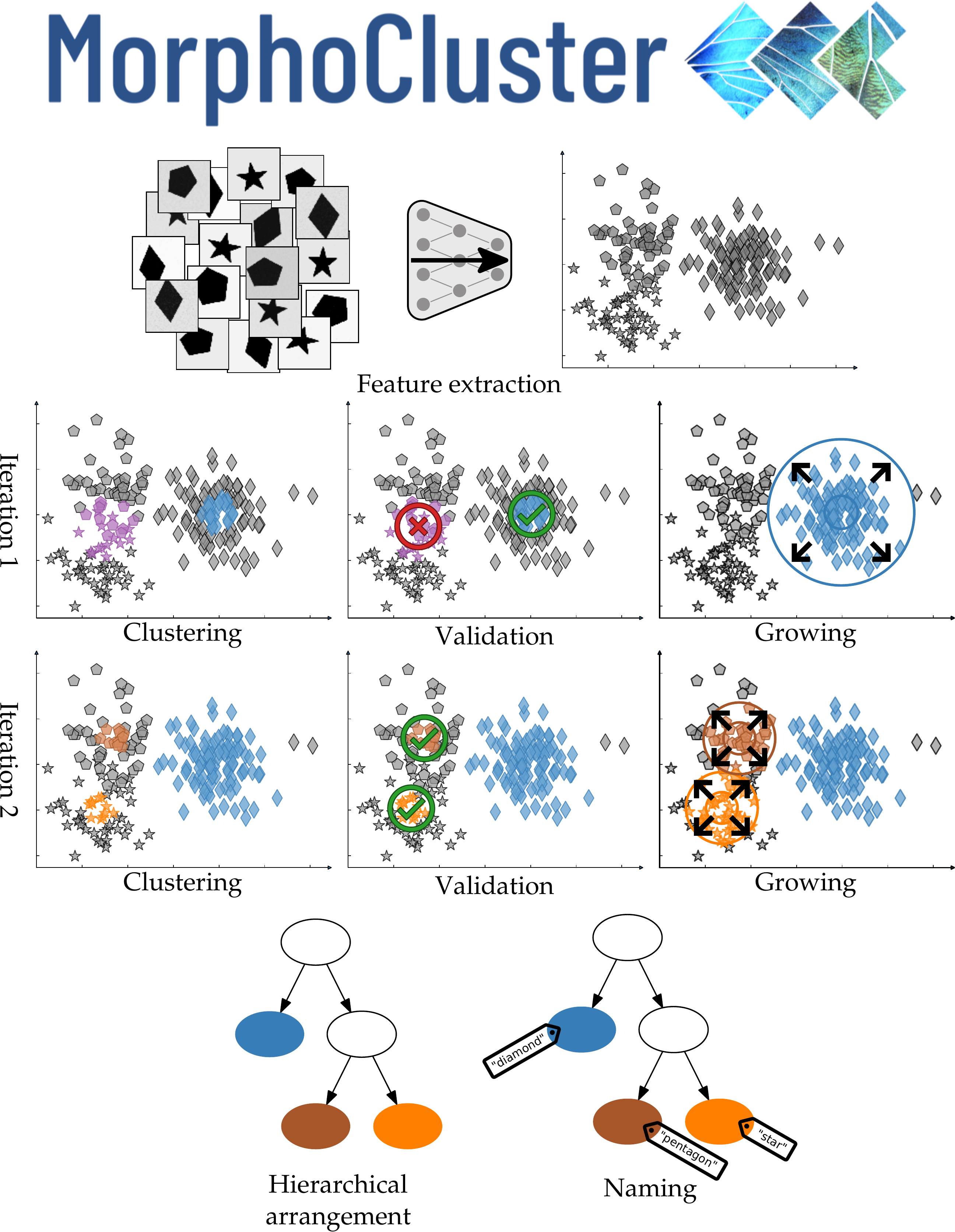}
	\caption{Overview of the MorphoCluster method.
	Images are projected to the feature space.
	Iteration~1: The blue cluster is validated and grown, the purple one rejected.
	Iteration~2: The orange and brown clusters are validated and grown.
	Finally, the clusters are arranged in a hierarchy and named.}
	\label{fig:method-overview}
\end{figure}

The MorphoCluster process is outlined in \cref{fig:method-overview}.
First, a deep feature extractor is trained to obtain features that encode relevant object properties for the task of plankton recognition and therefore enable efficient clustering.
Then, the entire data set is clustered using HDBSCAN* with settings that allow for the creation of large and homogeneous clusters.
In the \emph{cluster approval} phase, visually pure clusters are validated and mixed clusters are rejected manually.
During \emph{cluster growing}, the remaining pure clusters are used as seeds to find additional visually similar objects. The samples that are not assigned to a cluster after the growing step are re-clustered with a less restrictive setting that yields smaller clusters than in the previous round. Cluster approval and growth steps are thereafter repeated. The described process is conducted iteratively with less and less restrictive settings until no further meaningful clusters are found.
Thereafter, the identified clusters are hierarchically arranged using agglomerative clustering to group similar clusters. The clusters and branches of the resulting tree can then be inspected manually, very similar clusters can be merged and clusters and branches named in a user-defined manner.
Validation, growing and naming are conducted in a specially developed web application available at https://github.com/morphocluster.

\subsection{Data set used}%
\label{subsec:data-set}

We evaluate our approach on a data set~\cite{Kiko2020} of readily segmented grayscale images of individual particles in the water column which were acquired using the Underwater Vision Profiler 5 (UVP5)~\cite{Picheral2010}. The depicted objects are very small (\SI{100}{\micro\meter} to several centimeters) and their orientation is unrestricted.
The data set contains 1M unlabeled images and 584k labeled images that were sorted by experts into a selection of 65 classes
from a taxonomy based on the widely recognized WoRMS~\cite{Costello2013} taxonomy using EcoTaxa.
In that, the data set is similar to the ZooScanNet data set~\cite{Elineau2018}.

We call the initially \emph{\textbf{u}nlabeled} set of images \(\mathcal{U}\), the initially \emph{\textbf{l}abeled} set \(\mathcal{L}_0\).
The labeled data shows a severe class imbalance; the 10\% most populated classes contain more than 80\% of the objects and the class sizes span four orders of magnitude.

Like \cite{Orenstein2017} and \cite{Malde2019b}, we assume that the training set will be sufficient to learn features suitable for the distinction of all known and novel categories alike and that the distance in the feature space between two objects serves as a proxy for their similarity.
To evaluate the ability of MorphoCluster to detect novel classes, we select four \emph{\textbf{i}ndicator classes} \(\mathcal{C}_\text{i}\) (Veliger, Poeobius, T001, Flota) that are not used in the supervised training step.

The labeled set \(\mathcal{L}_0\) is split into a \emph{\textbf{t}raining set} \(\mathcal{L}_t\) of 392k objects and a \emph{\textbf{v}alidation set} \(\mathcal{L}_v\) of 192k objects, stratified by class.
\(\mathcal{L}_t\), without the indicator classes \(\mathcal{C}_\text{i}\), is used to train the feature extractor.
\(\mathcal{L}_v\) is first used to monitor the feature extractor training (ignoring \(\mathcal{C}_\text{i}\)) and later to evaluate the main MorphoCluster sorting process (including \(\mathcal{C}_\text{i}\)).

To validate the outcome of the MorphoCluster progress, we combine \(\mathcal{L}_v\) and \(\mathcal{U}\) and sort them jointly.
\(\mathcal{L}_v\) enables us to map the categories annotated with MorphoCluster to the annotations made with EcoTaxa.
The included indicator classes \(\mathcal{C}_\text{i}\) enable us to check if the MorphoCluster process allows detecting novel classes that the feature extractor was not trained on.

\subsection{Supervised training and feature extraction}%
\label{subsec:train-extract}

The supervised training of the feature extractor is a preliminary step to acquire knowledge about the discriminative features of the data at hand.
Transfer learning~\cite{Oquab2014} allows the reuse of information provided by labeled samples to obtain features that are actually relevant to taxon identification.

The images of the training and validation sets \(\mathcal{L}_t\) and \(\mathcal{L}_v\) (excluding the indicator classes \(\mathcal{C}_\text{i}\)) are used to train the network and monitor the classification loss, respectively.
We select a ResNet18~\cite{He2015b} as the backbone of the feature extractor as it shows a favorable accuracy-speed trade-off compared to other network architectures~\cite{Canziani2016}.
The network is initialized with weights pre-trained on the ImageNet data set~\cite{Deng2009} and fine-tuned to the classification task at hand following the common practice~\cite{Chatfield2014}.
To counter the class imbalance in the training set, we randomly sample up to 250 images from each class for each training epoch independently.
Early stopping is used to avoid overfitting.
The initial learning rate is set to \(1 \times 10^{-4}\) and decreased whenever the validation loss (measured on \(\mathcal{L}_v\)) plateaus until it reaches \(1\times10^{-8}\).
To consider all classes equally, we weight the validation loss by the inverse class size.
The batch size is set to 128 images.
The images are cropped to their tight bounding box and padded to a square with a minimum edge length of 128px.
Images larger than 128px are shrunken to this size.
The gray values are scaled to the \([0, 1]\) range.
We perform training-time augmentation using random rotations in \SI{90}{\degree} steps, random horizontal and vertical flips and additive Gaussian noise with
\(\sigma = 0.001\).
The models are trained using the PyTorch deep learning library~\cite{Paszke2017} on a NVIDIA GeForce GTX 1070 GPU.

Originally, the ResNet18 network produces a 512d feature vector for each image.
In a fine-tuning step, an additional layer is trained to reduce the number of features to 32 to reduce computation time and storage requirements in the subsequent steps.

After removing the classifier layer, the decapitated network serves as a feature extractor.
It is used to calculate feature vectors for all images in the data set (including labeled and unlabeled images).

\subsection{Clustering}%
\label{subsec:clustering}

In this completely unsupervised stage, the images of the unlabeled set \(\mathcal{U}\) and the validation set \(\mathcal{L}_v\) (including the \enquote{novel} indicator categories \(\mathcal{C}_\text{i}\)) are clustered jointly according to their feature vectors generated in the previous step.

To accumulate similar objects, we use the hierarchical density-based HDBSCAN* algorithm~\cite{McInnes2017,Campello2015} which has some favorable properties: It handles clusters of variable density, makes few assumptions about the data distribution, has a small number of parameters, and is robust to outliers.
Another remarkable property is that HDBSCAN* clusters only the most dense regions of the feature space and rejects most of the objects as noise.
This is favorable in our setting, since this way, the resulting clusters are very pure.

HDBSCAN* is parameterized by the neighborhood size \(k\) and the minimum cluster size \(m\).
We set neighborhood size \(k=1\) and vary minimum cluster size \(m\) throughout the iterations of validation and growing.
Initially, a large value is chosen for \(m\) to extract the largest coherent groups first.
It is decreased after each iteration of the process so that increasingly smaller clusters are found.
This strategy is used to remove large groups of similar objects early in the process and to keep the number of clusters to be validated and grown in each step small.
Too small values for \(m\) would lead to excessive fragmentation of the data resulting in many small clusters leading to a drastically increased effort in the following steps.

The detected dense regions of the feature space serve as \emph{cluster seeds} for the further treatment in the following steps.

\subsection{Cluster validation}%
\label{subsec:validation}

\begin{figure}
	\includegraphics[width=0.75\textwidth]{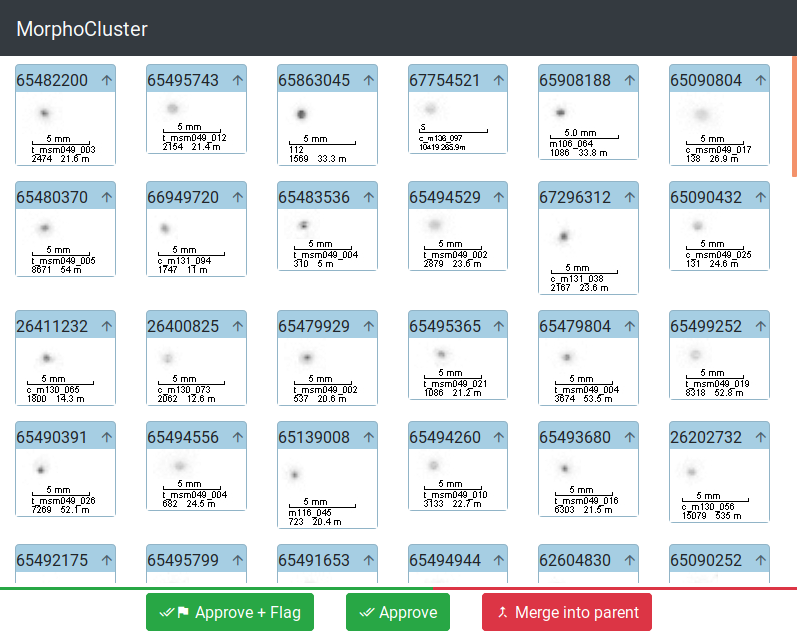}
	\caption{User interface for cluster validation.
		The images of a cluster are presented to the user.
		\enquote{Approve} marks a cluster as
		validated (=being pure). \enquote{Approve + Flag} additionally flags the
		cluster for preferred treatment during the growth step. \enquote{Merge into parent} deletes a cluster
		and moves its objects back to the pool of unclustered objects. Above the
		buttons, a progress indicator is visible.}
	\label{fig:validate}
\end{figure}

\Cref{fig:validate} shows the user interface for manual cluster seed validation and review.
One after the other, each cluster seed is displayed to the user.
Its images are arranged in an alternating fashion so that two neighboring images are maximally dissimilar with respect to their deep learning features. The resulting contrast facilitates the annotator's judgment.
The user then flags homogeneous cluster seeds as \enquote{validated}.
Impure cluster seeds are deleted and the corresponding objects are returned to the pool of unclustered objects.

\subsection{Cluster growing}%
\label{subsec:subgrowing}

After validation, only pure cluster seeds are left.
Due to their construction (see \cref{subsec:clustering}), a seed is only the very core of a dense region.
The purpose of \emph{cluster growing} is therefore the accretion of further images from the neighborhood of this dense region until the boundaries of a cluster are reached.

\begin{figure}
	\centering
	\includegraphics[width=0.75\textwidth]{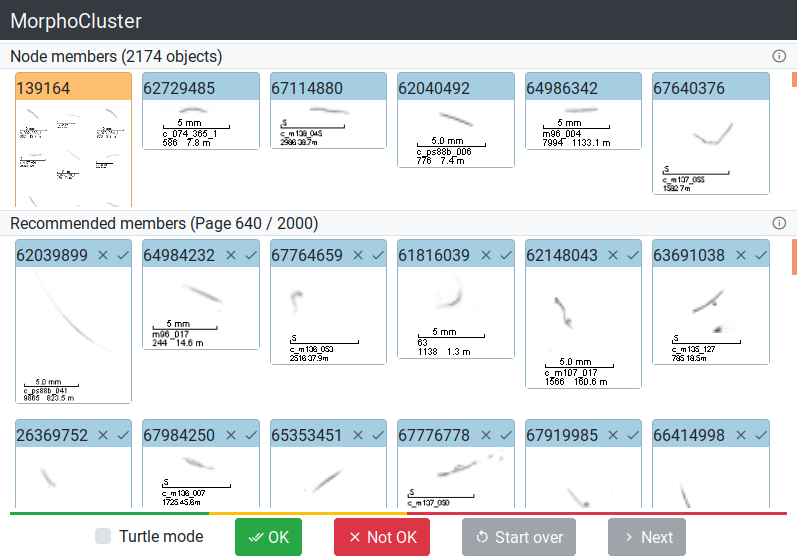}
	\caption{User interface for growing clusters. The top half of the screen displays
		the current member objects of a cluster. The bottom half always shows a
		page of 50 candidate members.\\
		The colored bar above the buttons visualizes the search interval for the
		pages of candidates that should be added to the cluster. Pages in green were
		judged to match the cluster, pages in red were judged not to match.
		Pages in yellow were not reviewed yet.}
	\label{fig:screenshot-grow}
\end{figure}

For each cluster, the objects that make up the cluster seed are presented to the user (\Cref{fig:screenshot-grow}).
The objects that are so far no member of any cluster are displayed as \emph{recommended members} ordered by decreasing similarity to the cluster seed (measured by their distance to the seed's centroid).
The user then needs to find the first object in the list of recommended members that is not similar to the seed images.
Finally, the objects earlier in the list (being more similar) are added to the cluster.
This setup is similar to the visual search engine in \cite{Joly2014}.
The list of recommended members is partitioned into pages of 50 objects that are reviewed jointly.

The application assists in finding the similarity threshold by employing binary search to minimize the number of objects that a user has to review.
In the first stage of the task, the right limit of the search interval (a point where all objects are strictly dissimilar) is determined:
Beginning with the first page, the images of selected pages are reviewed if they match the seed images.
The number of pages that are skipped between successive page reviews is doubled in each step.
If the images start to differ from the seed images, the right limit of the search interval for the cluster radius is found.

Subsequently, the actual binary search step narrows down the search interval to find the last page with matching candidate objects.
Because many objects are never seen by the user, the process is much faster than adding each object to the cluster individually.

This approach is permitted under the assumption that if all objects on a certain page are sufficiently similar to
the seed, all objects of the previous pages are also similar to the seed.

A so-called \enquote{turtle mode} allows for a very detailed examination and definition of the cluster border
by allowing single objects to be removed from the set of recommended members.
Once an individual object is removed from the current page, turtle mode is activated and binary search is disabled.
Now, in turtle mode, all remaining objects have to be validated individually and the speed-up provided by binary search is traded for accuracy.

\subsection{Cluster naming}%
\label{subsec:naming}

After the objects are treated and moved to clusters, these clusters are named with computer-assistance using the respective function of the MorphoCluster application.
To this end, the list of clusters is transformed into a hierarchy by agglomerative clustering of the cluster centroids using average linkage (UPGMA) clustering~\cite[p.~76]{Everitt2011}.
The resulting automatic hierarchy serves as a starting point for a user-defined taxonomy.
Arranging clusters in a hierarchy makes them easier to annotate because many of the clusters found in the previous steps are very similar and can be given the same name or fall into the same superclass.
Their similarity in the feature space makes them close neighbors in the thus defined tree.
The tree is presented to the annotator, who can merge clusters if they are perceived as being identical.
The annotator can also rearrange individual nodes and give them names.
To this end, we started at the leaves of the tree and worked our way up to the root.
Whenever a node looked different than its siblings, it was given a distinct name and moved up in the hierarchy.
In the end, the name of each node was transferred to its corresponding objects.
The resulting set of now labeled images is called \(\mathcal{L}_{MC}\).

\subsection{Experimental approach}%
\label{subsec:exp_approach}
We applied the entire process of clustering, cluster approval, cluster growth and naming to the combination of images from the unlabeled set \(\mathcal{U}\) and the  validation set \(\mathcal{L}_v\) (including the indicator classes \(\mathcal{C}_\text{i}\)). Annotator actions were tracked during the approval, growth and naming steps to monitor the time spent during each step. To account for longer breaks, the log was split into \emph{sessions} that contained no breaks longer than ten minutes.
The duration of a session is the time span between its first and last entry.

For the evaluation of their precision, up to 500 objects per class\footnote{Some classes are smaller.} were randomly sampled from \(\mathcal{L}_v\),
for \(\mathcal{L}_{MC}\) only 400, due to the larger number of classes.
The samples of each class were manually reviewed and outliers (false positives) were removed.
The precision of a category is then the fraction of inliers.

The precision of \(\mathcal{L}_{MC}\) and \(\mathcal{L}_0\) in this analysis is a measure of self-consistency because the same person (R. Kiko) that did the sorting in MorphoCluster and in large parts that of the initial data set also evaluated the sub-samples.

\subsection{Evaluation metrics}
The \emph{precision} of a class $c$ is the number of objects \emph{correctly} classified as $c$ (true positives) divided by the total number of objects classified as $c$ (true positives and false positives):
\[
	Pr_c = \frac{TP_c}{TP_c + FP_c}
\]

\emph{Macro precision} is the arithmetic mean of all individual precisions:
\[
	\overline{Pr} = \mean\limits_{c}Pr_c
\]

Given two different labelings \(\mathcal{L}_a\) and \(\mathcal{L}_b\) of the same objects,
we define the \emph{relative overlap} of two classes $c_a$ from \(\mathcal{L}_a\) and $c_b$ from \(\mathcal{L}_b\)
as the number of objects that are assigned to \emph{both} $c_a$ and $c_b$ divided by the number of objects assigned to \emph{either} of them:
\[
	RelOverlap = \frac{\left|c_a \cap c_b\right|}{\left|c_a \cup c_b\right|} = \frac{\left|c_a \cap c_b\right|}{\left|c_a\right| + \left|c_b\right| - \left|c_a \cap c_b\right|}
\]

\FloatBarrier

\section{Results}%
\label{sec:experiments-results}

\subsection{Supervised training}%
\label{subsec:exp:supervised-training}

\begin{table}[tp]
	\sisetup{table-format=1.3}
	\begin{tabular}{lSS}
		\toprule
		{}              & {512d} & {32d}    \\
		\midrule
		Accuracy        & 0.560797                  & 0.557159 \\
		Macro Precision & 0.294231                  & 0.301738 \\
		\bottomrule
	\end{tabular}
	\caption{Accuracy and precision of the classifier trained for feature extraction before (512d) and after dimensionality reduction (32d).
		Dimensionality reduction did not substantially change the capacity of the classifier.}
	\label{tab:res-supervised}
\end{table}

\begin{figure}[tp]
	\includegraphics[width=\textwidth]{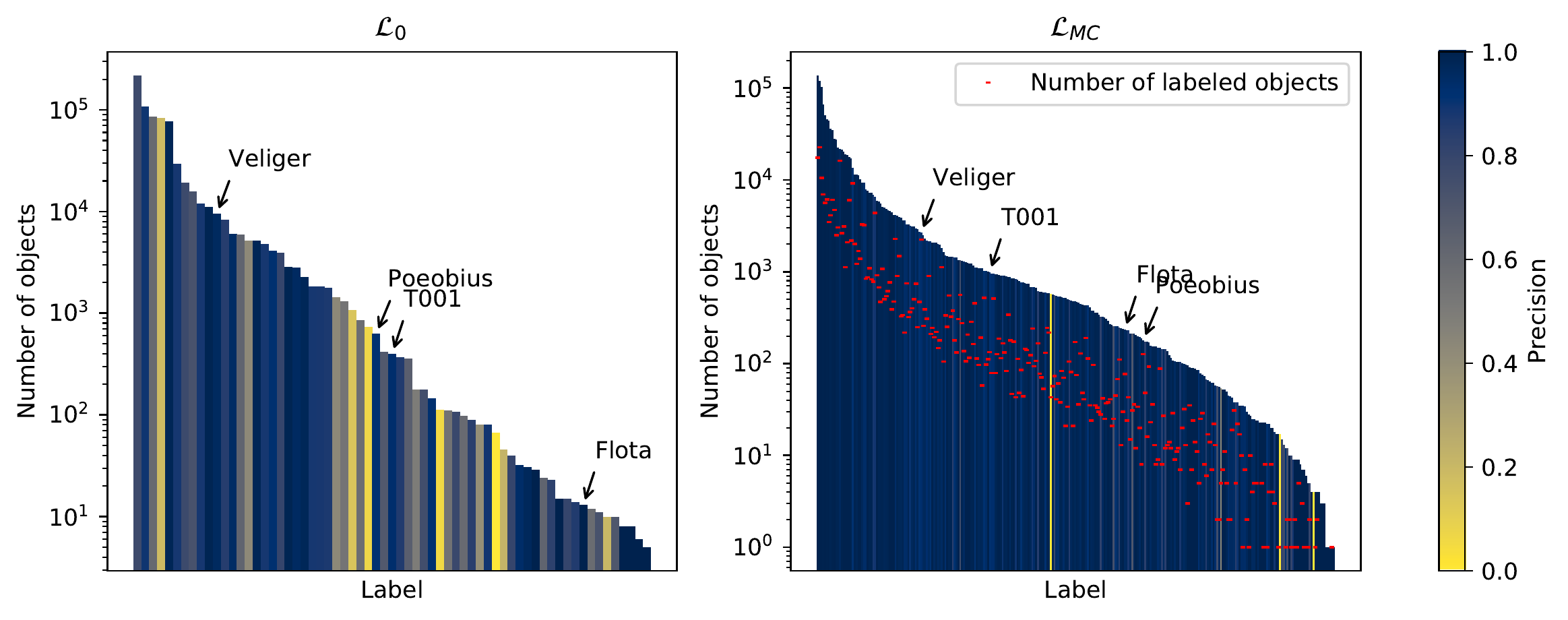}
	\caption{Classes in the initial labeling (\(\mathcal{L}_0\))
		and the MorphoCluster labeling (\(\mathcal{L}_{MC}\)) ordered by their size.
		Four indicator classes \(\mathcal{C}_\text{i}\) (Veliger,
		Poeobius, T001, Flota), indicated by arrows, are used to evaluate the ability of MorphoCluster to detect
		novel classes. The class sizes of \(\mathcal{L}_0\) and \(\mathcal{L}_{MC}\)
		are in the same range, but the latter contains many more classes.
		The precision of each class is color-coded (see \cref{subsec:accuracy}).
		The number of objects from \(\mathcal{L}_0\) in each class of \(\mathcal{L}_{MC}\)
		is denoted in red. It is roughly one order of magnitude lower than the MorphoCluster class size.
	}
	\label{fig:abundance}
\end{figure}

The trained classifier achieved comparatively low scores even when using the full set of 512 feature dimensions (\cref{tab:res-supervised}).
This could be expected as the overall macro precision of the training set \(\mathcal{L}_0\) was also only \num{0.738}, with some classes showing very low precision (\cref{fig:abundance}; left).
The feature reduction to 32 dimensions did not compromise classification performance substantially and even increased macro precision by a small amount (\cref{tab:res-supervised}).
We did not optimize the hyper-parameters of the network for high classification scores to maintain its generalization capabilities as a feature extractor.

\subsection{MorphoCluster efficiency}%
\label{subsec:efficiency}

\begin{figure}
	\includegraphics[width=\textwidth]{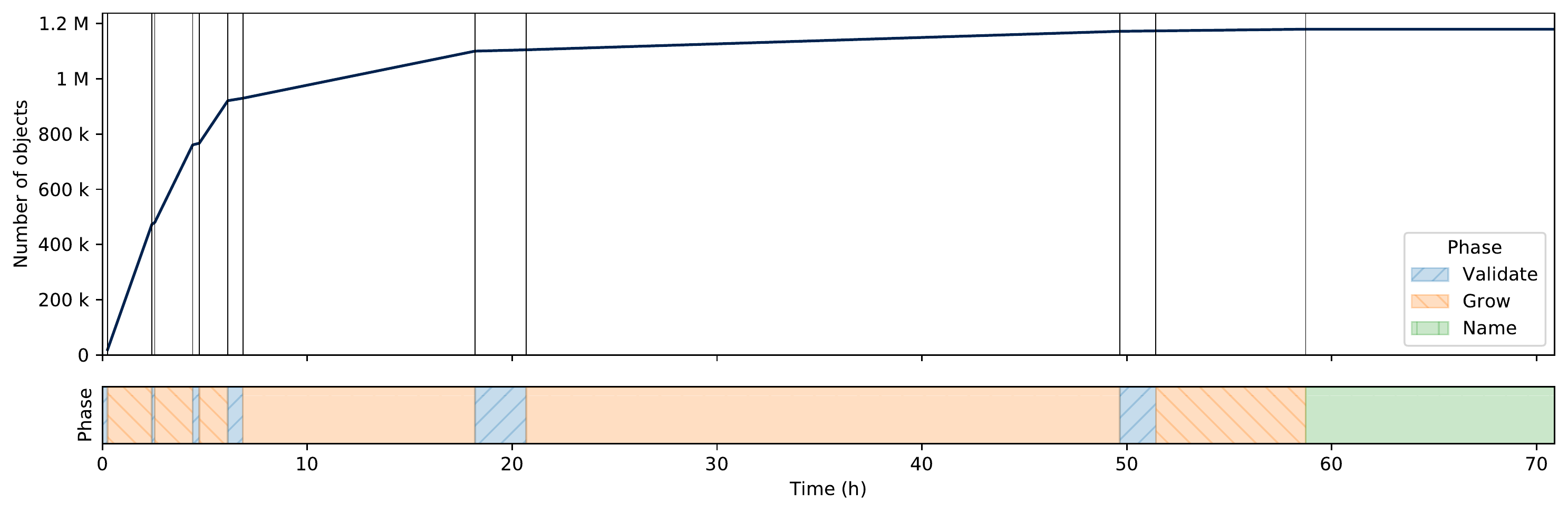}
	\caption{Number of validated objects during data annotation.
	The time periods are colored according to their respective phase.
	}
	\label{fig:project-steps}
\end{figure}

The metrics collected during the iterative cluster validation and growing steps of the MorphoCluster process are depicted in \cref{tab:obj_h} and \Cref{fig:project-steps}.
The number of clusters found in each iteration increased as a function of the minimum cluster size \(m\).
Most of the proposed cluster seeds were validated which indicates that the calculated clusters are in fact very pure.
Only a few objects were assigned to clusters during the validation phases because the cluster seeds consist only of the densest regions.
Growing a cluster added a large number of objects from the neighborhood of a cluster and the majority of objects were assigned to clusters during growing.
During the first rounds of validation and growing, very large clusters were identified that mainly contained detritus-like objects.
During later rounds, smaller clusters containing more rare objects (e.g.\ copepods, veliger larvae etc.) were validated and grown.
\Cref{fig:project-steps} shows the number of objects sorted per hour during the entire MorphoCluster process.
Most time was spent in the validation and growing steps to group similar parts of the data set and assignment of names to the identified clusters only accounts for a fraction of the total time.
Validation and growing alone took \SI[round-mode=figures]{\constTotalDurationValGrowH}{\hour}. \num[round-precision=0]{\constValGrowObjPH} objects were sorted per hour when considering these steps in isolation.
Naming took \SI[round-mode=figures]{\constTotalDurationNamingH}{\hour}.
The first three rounds of validation and growing yielded remarkably high sorting speeds (\cref{fig:project-steps}).
After that, sorting got drastically slower in each iteration.

\begin{table}
	\input{project_phases_simple.tex}
	\caption{
		Iterations in the MorphoCluster process with metrics in each step.
		Minimum cluster size \(m\); Number of proposed new clusters; Number of validated clusters;
		Number of objects sorted per hour.
		Note that at this point, the raw clusters have not been grouped and named yet.
	}
	\label{tab:obj_h}
\end{table}

\subsection{Hierarchical ordering and naming}

\begin{figure}
	\begin{minipage}[c]{.33\textwidth}
		\hspace*{0.03\textwidth}%
		\includegraphics[width=0.94\textwidth]{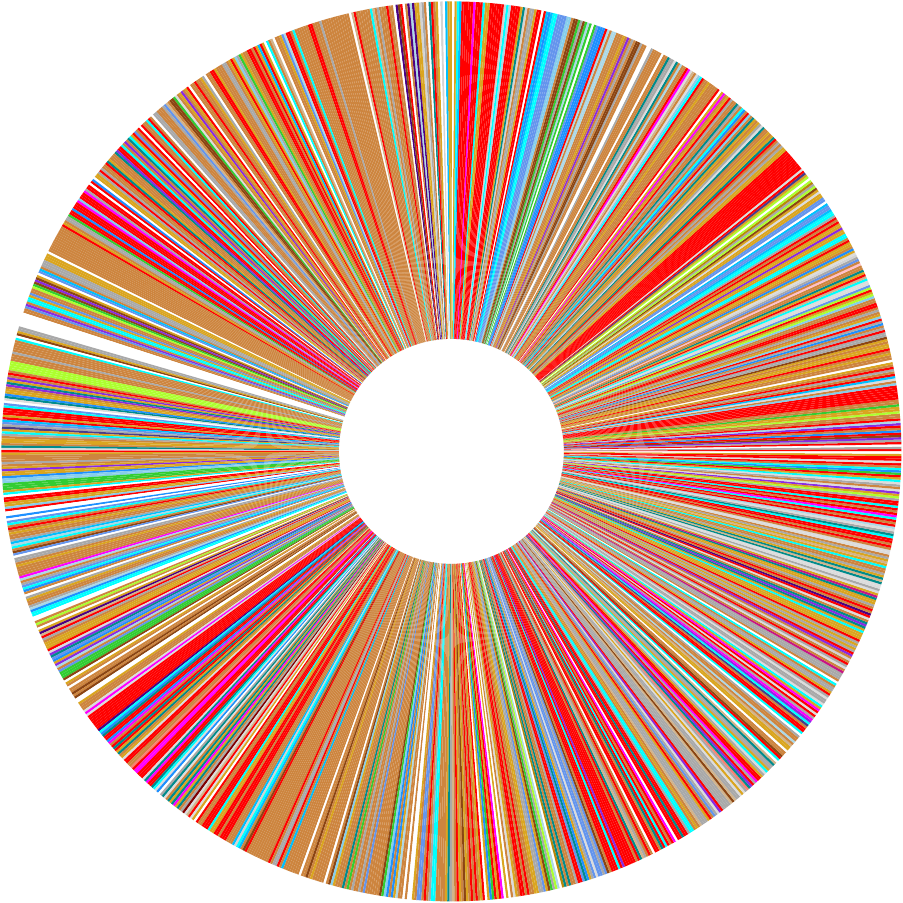}
	\end{minipage}%
	\begin{minipage}[c]{.33\textwidth}
		\hspace*{0.03\textwidth}%
		\includegraphics[width=0.94\textwidth]{tree_orig.pdf}
	\end{minipage}%
	\begin{minipage}[c]{.33\textwidth}
		\hspace*{0.03\textwidth}%
		\includegraphics[width=0.94\textwidth]{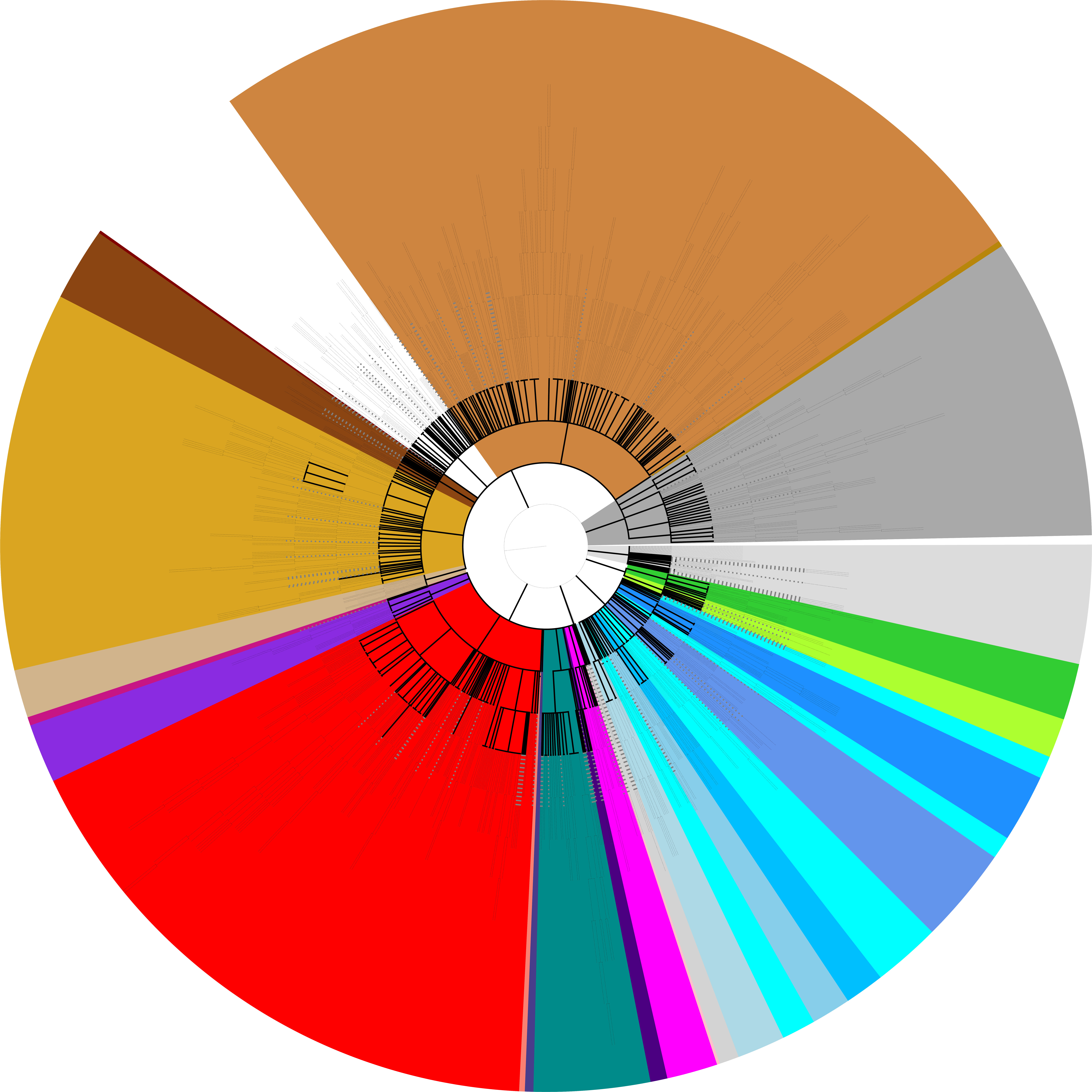}
	\end{minipage}
	\includegraphics[width=0.9\textwidth]{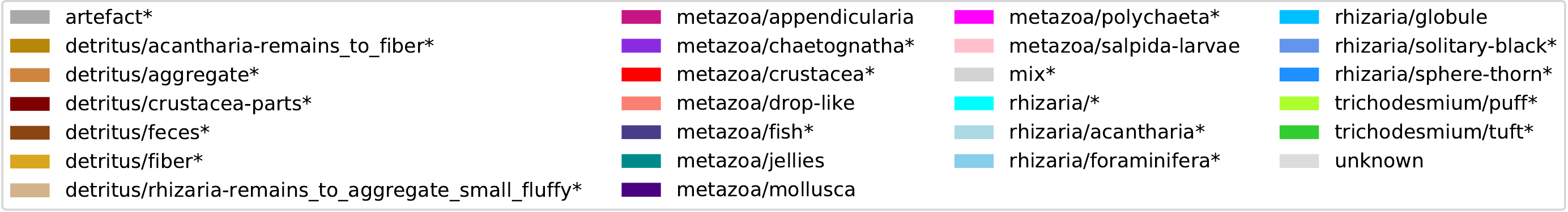}
	\caption{Unordered nodes (left), automatic hierarchy (middle) and revised hierarchy with named branches denoted by bold lines (right).
	Corresponding sections of the three charts are colored alike according to broad supercategories.}
	\label{fig:trees}
\end{figure}

\Cref{fig:trees} displays the \num{1192} unordered clusters as the result of the iterated clustering, approval and growing (left),
their automatic hierarchical organization (middle) with \num{2382} nodes and the revised hierarchy after reordering and naming (right) with \num{280} named branches (bold) in \num{26} broad supercategories (colored).
It is apparent that the initial hierarchy already introduced a high level of order and contained large branches
that were pure with respect to the considered supercategories.
However, branches that belong to the same supercategory according to expert knowledge were still scattered throughout the tree.
To obtain the final result (right), these branches were manually mounted to a common supercategory and
relevant branching points were named using free-form input.
This also reduced the depth of the tree from 23 to 12.
The final result illustrates yet again that these supercategories are finely branched.
Re-arranging the initial hierarchy and naming the branches took \SI[round-mode=figures]{\constTotalDurationNamingH}{\hour},
only \SI[round-mode=figures]{\constTotalDurationNamingPercent}{\percent} of the total time.

Considering this step in isolation, \num[round-precision=0]{\constNamingObjPH} objects or \num[round-precision=0]{\constNamingClassesPH} complete classes were labeled per hour.
Including validation, growing and naming, we spent a total \SI[round-mode=figures]{70.88}{\hour} on sorting \num{1179619} objects into a set of \num{280} new categories (\num{16641} objects per hour) while most objects were already sorted in the early steps.

\subsection{Completeness}
\num{16400} (\SI[round-mode=figures]{1.3712156746673758}{\percent}) \emph{residual objects} were not assigned using the MorphoCluster approach
because they were neither clustered and validated nor moved to an existing cluster in the growing step.
They were ultimately left untreated.

58 of the 65 classes in the initial labeling \(\mathcal{L}_0\) were reproduced in the new labeling \(\mathcal{L}_{MC}\),
while objects from some initial classes (\input{empty_correspondences.tex}) could not be reproduced.
In part, their objects were not put into any class at all, in part their objects were included in other classes.
All of these categories contain less than 40 objects and/or show high intra-class variability.
Moreover, images of \emph{Pyrosomatida\_Pyrosoma} (large colonies of individual animals) are very
large and down-scaling them to the fixed input size of the feature extractor network removes nearly
all of their distinctive features.

\subsection{Accuracy}%
\label{subsec:accuracy}

\begin{table}
	\sisetup{table-format=1.3}
	\begin{tabular}{lSSScSSc}
		\toprule
		                                         & \(\mathcal{L}_{MC}\) vs \(\mathcal{L}_v\) &
		\multicolumn{3}{c}{\(\mathcal{L}_{MC}\)} & \multicolumn{3}{c}{\(\mathcal{L}_0\)}
		\tabularnewline
		\cmidrule(lr){3-5}\cmidrule(lr){6-8}
		                                         & {\(\overline{Pr}\)}
		                                         & {\(\overline{Pr}\)}                                  &
		{Pr$_{10}$}                              & {N}                                   &
		{\(\overline{Pr}\)}                                     &
		{Pr$_{10}$}                              & {N} \tabularnewline
		\midrule
		Total                                    & 0.650
		                                         & 0.948963                              &
		0.888889                                 & 280
		                                         & 0.738
		                                         & 0.288459                              &
		65\tabularnewline
		Living                                   & 0.719
		                                         & 0.947178                              &
		0.879729                                 & 126
		                                         & 0.644
		                                         & 0.187400                              &
		42\tabularnewline
		Non-living                               & 0.592
		                                         & 0.952001                              &
		0.935000                                 & 146
		                                         & 0.862
		                                         & 0.730000                              &
		11\tabularnewline
		\bottomrule
	\end{tabular}
	\caption{
		Comparison of precision.
		The columns show the macro precision of MorphoCluster according to the original labels (\(\mathcal{L}_{MC}\)
		vs \(\mathcal{L}_v\)) and the macro precision of \(\mathcal{L}_{MC}\) and \(\mathcal{L}_0\) according to manual examination.
		\(\overline{Pr}\) is the macro precision, Pr$_{10}$ the \SI{10}{\percent} quantile of individual precisions, and N the number of classes.
		The results are further broken down by living (animals, plants) and non-living (fibers, aggregates, feces,~\dots) categories.
	}
	\label{tab:precision}
\end{table}

Using MorphoCluster, a very large fraction of classes was sorted with high precision.
\Cref{fig:abundance} shows the class size and individual precision per class which is consistently higher for \(\mathcal{L}_{MC}\) compared to \(\mathcal{L}_{0}\). 
Roughly a tenth of the objects in each class in \(\mathcal{L}_{MC}\) was already labeled in \(\mathcal{L}_0\) (red) which allows calculating the agreement between both labelings.
\Cref{tab:precision} shows this agreement (\(\mathcal{L}_{MC}\) vs \(\mathcal{L}_v\)) and also the macro precision of \(\mathcal{L}_{MC}\) and \(\mathcal{L}_0\) individually.

For the calculation of the agreement between the MorphoCluster labeling \(\mathcal{L}_{MC}\) and the initial labeling \(\mathcal{L}_0\),
only \(\mathcal{L}_v\) was used to avoid overly optimistic results coming from data which the feature extractor was trained on.
We computed the proportions of objects from all initial classes in \(\mathcal{L}_v\) for every MorphoCluster category in \(\mathcal{L}_{MC}\).
Each category in \(\mathcal{L}_{MC}\) was then assigned its predominant \(\mathcal{L}_v\)-class-label.
The agreement was measured as the precision of a \(\mathcal{L}_{MC}\) class according to the respective predominant \(\mathcal{L}_v\) class.

To some degree, the labeling of MorphoCluster is consistent with the initial one
(\cref{tab:precision}, \(\mathcal{L}_{MC}\) vs \(\mathcal{L}_v\) in the first row).
The agreement is, however, consistently lower than the precision of \(\mathcal{L}_{MC}\) according to manual examination.
This suggests that MorphoCluster categories often contain objects from multiple initial categories.
The reason becomes apparent when looking at the precision of the initial labeling \(\mathcal{L}_0\) (\cref{tab:precision}, \(\mathcal{L}_0\)):
Macro precision over all categories is only \num{0.737578}, with \SI{90}{\percent} of the classes having a precision of only \num{0.2884594739667204} or higher.
In contrast, the precision of the MorphoCluster labeling \(\mathcal{L}_{MC}\)
is excellent (\cref{tab:precision}, \(\mathcal{L}_{MC}\)): Macro precision over all categories is
\num{0.948963}, with \SI{90}{\percent} of the classes having a precision of \num{0.88888888} or higher.

The categories were also divided into living and non-living categories and macro precision was calculated for each group individually.
Some categories (\enquote{unknown\_*}, \enquote{mix\_*}) could not be assigned to either
living or non-living and are therefore not included in these results.
According to \cref{tab:precision}, non-living categories are sorted with higher precision than living categories
in both \(\mathcal{L}_{MC}\) and \(\mathcal{L}_0\),
so it might be easier to be self-consistent on the classification of non-living categories.

\subsection{Fine-grained data set exploration}%
\label{subsec:fine-grained}

\Cref{fig:abundance} compares the initial labeling \(\mathcal{L}_0\) to the resulting labeling \(\mathcal{L}_{MC}\).
Using MorphoCluster, the data set could be sorted into \num{280} categories in contrast to the initial 65 categories.
Also, the relative class abundances of the indicator classes \(\mathcal{C}_\text{i}\) were misestimated in the initial sorting.
The high ranking of Poeobius in \(\mathcal{L}_0\) likely originates from the high effort that was put into finding examples for this class
after it had been discovered~\cite{Christiansen2018}.

Although the largest part of the data set was sorted in the early steps (see \cref{subsec:efficiency}),
\Cref{fig:project-steps} shows that the later steps were nevertheless required to achieve this large number of categories.

\begin{figure}
	\includegraphics[width=\textwidth]{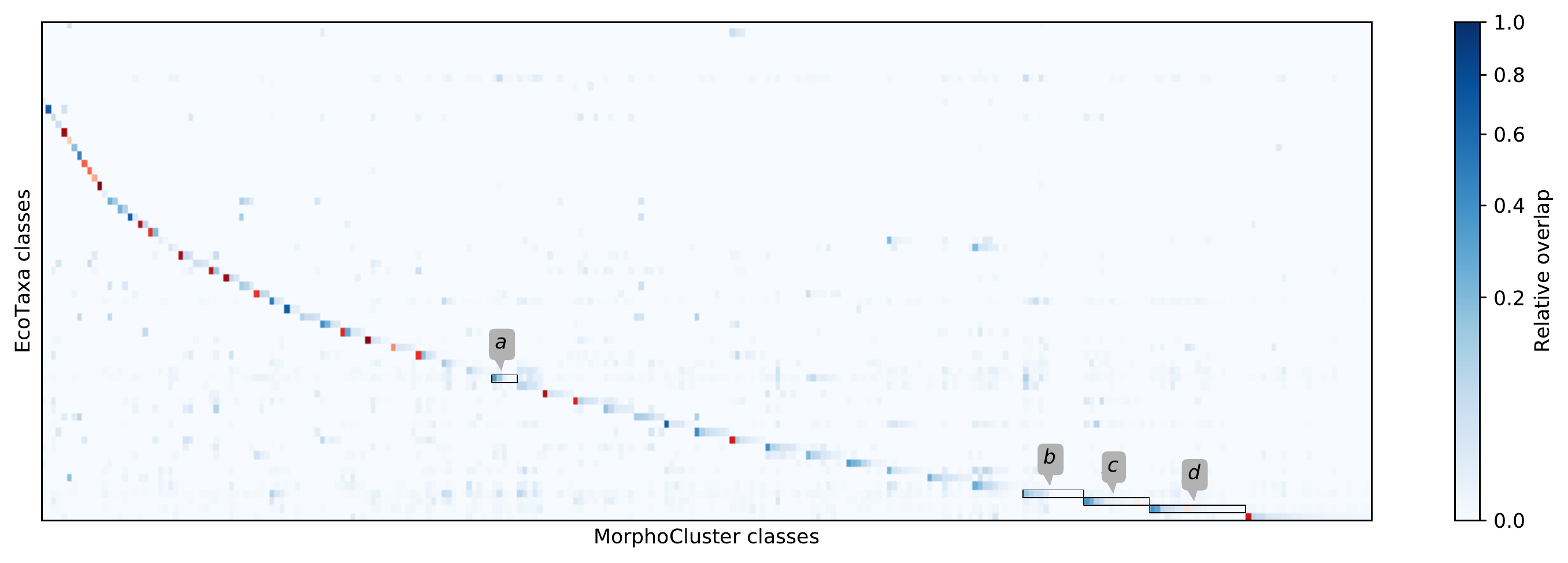}
	\caption{Correspondence of MorphoCluster and initial labels measured by relative overlap.
		\(\mathcal{L}_0\) classes are ordered by their number of correspondences,
		\(\mathcal{L}_{MC}\) classes are ordered by their corresponding \(\mathcal{L}_0\) class. Therefore a diagonal
		structure emerges.
		Manually established direct correspondences are colored using shades of red.
		The first rows are \(\mathcal{L}_0\) classes without a correspondence in \(\mathcal{L}_{MC}\).
		Selected \(\mathcal{L}_0\) classes are annotated for further analysis: $a$ fluffy\_light, $b$ fluffy\_dark, $c$ Trichodesmium\_puff,
		$d$ Maxillopoda\_Copepoda.
	}
	\label{fig:confusion-et-mc}
\end{figure}

\begin{figure}
	\includegraphics[width=\textwidth]{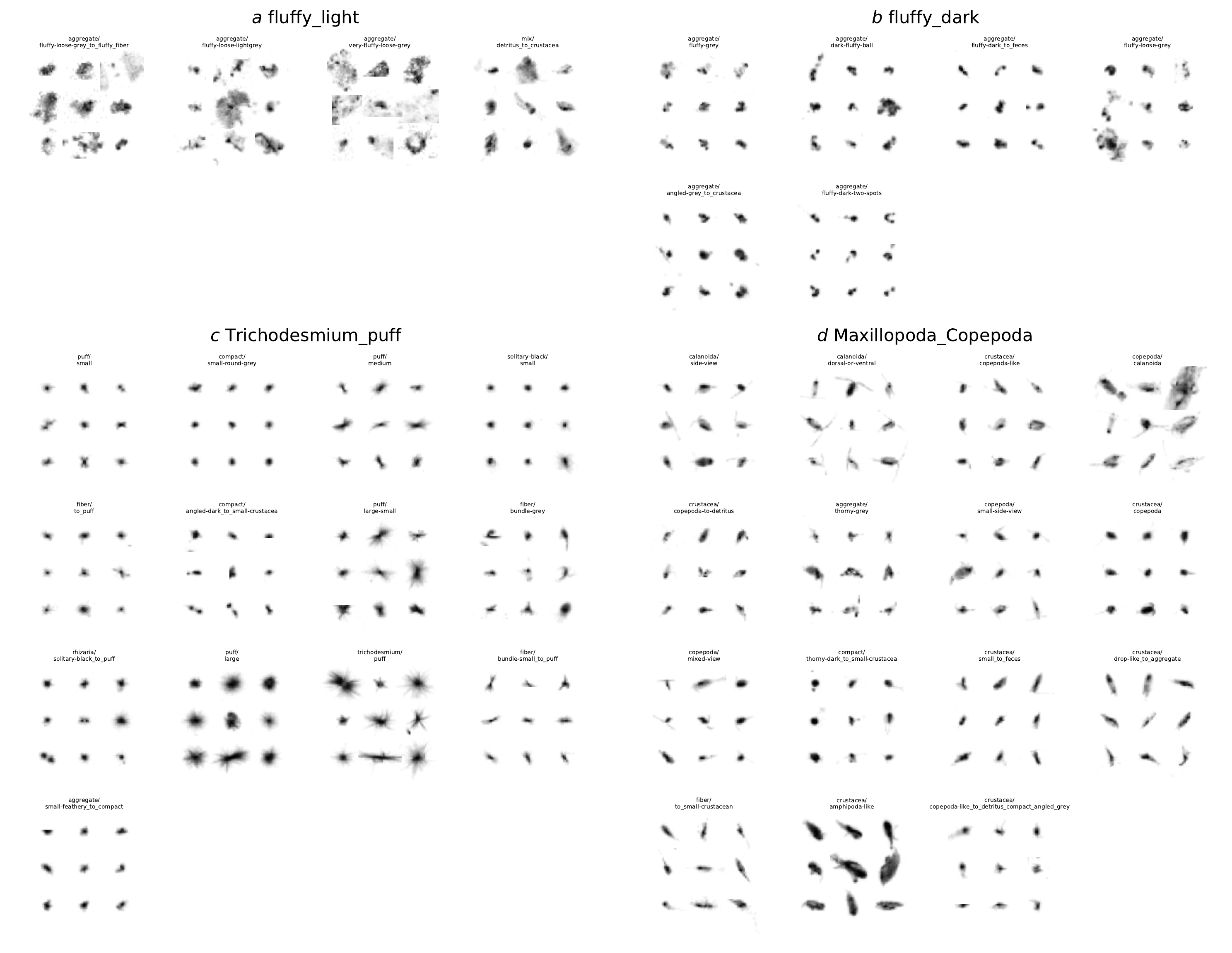}
	\caption{Four \(\mathcal{L}_0\) classes (denoted in \cref{fig:confusion-et-mc})
		and their corresponding \(\mathcal{L}_{MC}\) classes.
		These \(\mathcal{L}_0\) classes are highly diverse and can be split up into finer, very homogeneous groups
		using MorphoCluster.
	}
	\label{fig:split-up}
\end{figure}

\begin{figure}
	\includegraphics[width=\textwidth]{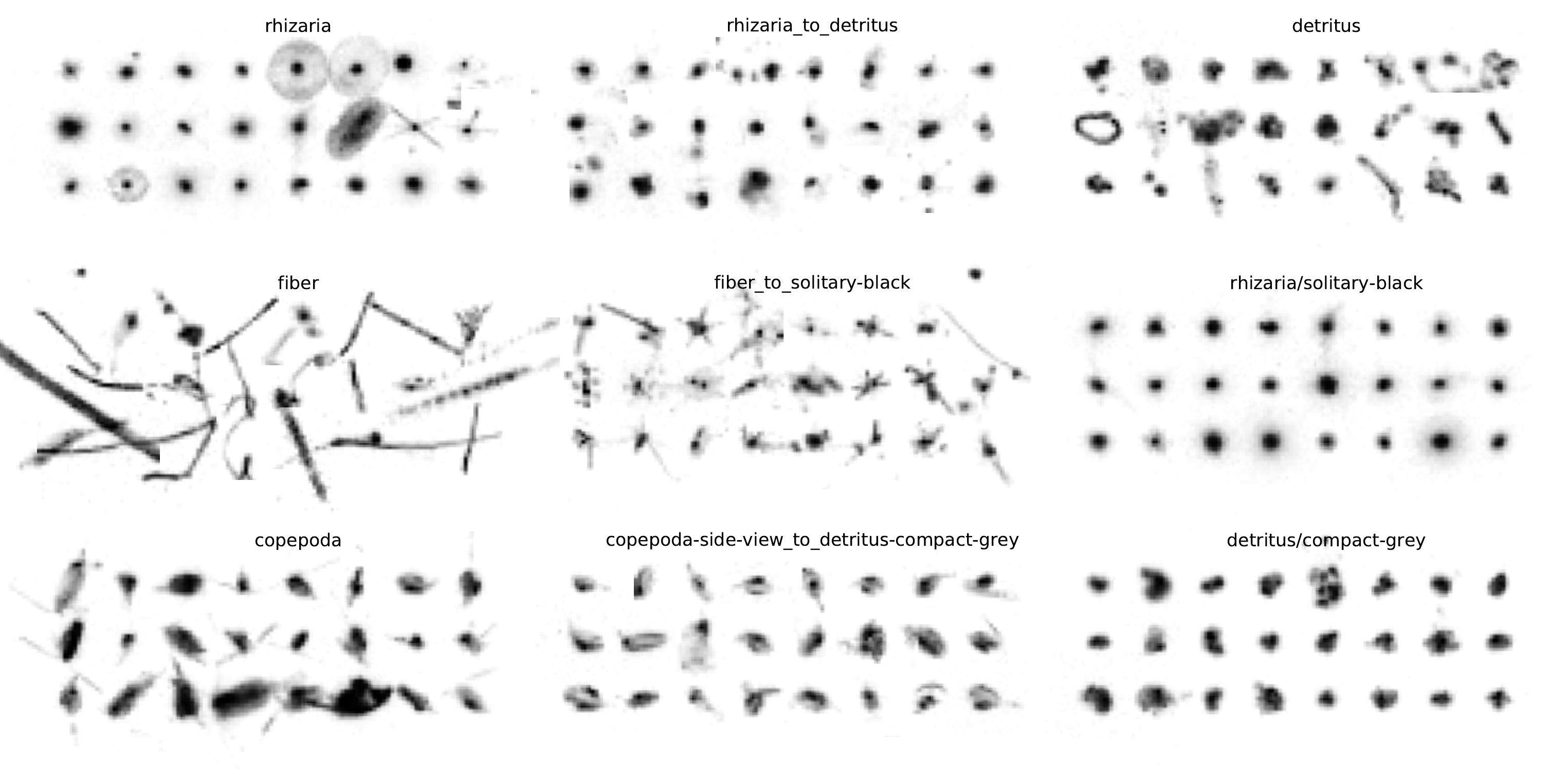}

	\caption{
		Some classes defined using MorphoCluster (\enquote{X\_to\_Y}, middle column) form a
		transition between two clear-cut classes.
	}
	\label{fig:transitional}
\end{figure}

Spiking the data with labeled objects from the validation set \(\mathcal{L}_v\) allowed the calculation of relative overlap between initial and new classes
\(\mathcal{L}_0\) and \(\mathcal{L}_{MC}\).
This relative overlap is depicted in the correspondence matrix \cref{fig:confusion-et-mc}.
For each \(\mathcal{L}_0\) class, the corresponding \(\mathcal{L}_{MC}\) classes are aligned by descending overlap in a horizontal group.
A single category in the initial labeling \(\mathcal{L}_0\) sometimes has a direct correspondence (red) and often decomposes into
multiple categories in the MorphoCluster labeling \(\mathcal{L}_{MC}\), partly into finer subcategories (entries in the same group),
partly into similar-looking but unrelated categories (entries elsewhere in the row).
Conversely, \(\mathcal{L}_{MC}\) classes often recruit their members from multiple \(\mathcal{L}_0\) classes, indicated by columns with multiple entries.
For a complete list of correspondences, see \cref{appendix:corresponding-labels}.

Subdivisions show that the images taken by the UVP5 could allow a more fine-grained sorting than previously attempted.
To illustrate the high level of diversity within the classes in the initial labeling
and the strong homogeneity within individual \(\mathcal{L}_{MC}\) classes,
the objects of four selected \(\mathcal{L}_0\) classes (annotated in \cref{fig:confusion-et-mc})
are depicted in detail in \cref{fig:split-up}.

Aggregations of objects from multiple original classes are signs that
the initial labeling was inconsistent or that the previously applied classification scheme did not fit the cluster structure in the data.
\(\mathcal{L}_{MC}\) also contains many transitional classes that lie in between two clear-cut
classes, as depicted in \cref{fig:transitional}.
These contain objects that can not be assigned to either of both categories with certainty.
In most cases, these seem to be decaying organisms that are losing their distinctive morphological
features and seem to turn into dead matter (detritus).
Some classes were annotated in \(\mathcal{L}_{MC}\) that did not share any objects with an existing
class in \(\mathcal{L}_0\), most of them being detritus subcategories. These are not included in the correspondence matrix.

In summary, these results suggest that the subdivisions, aggregations and transitional classes in \(\mathcal{L}_{MC}\) go beyond the previous labeling \(\mathcal{L}_0\) by refining it.
Decision boundaries seem to align better with the data structure.

\subsection{Novelty detection}%
\label{subsec:novelty-detection}

\begin{figure}
	\begin{minipage}[t]{.45\textwidth}
		\includegraphics[width=\textwidth]{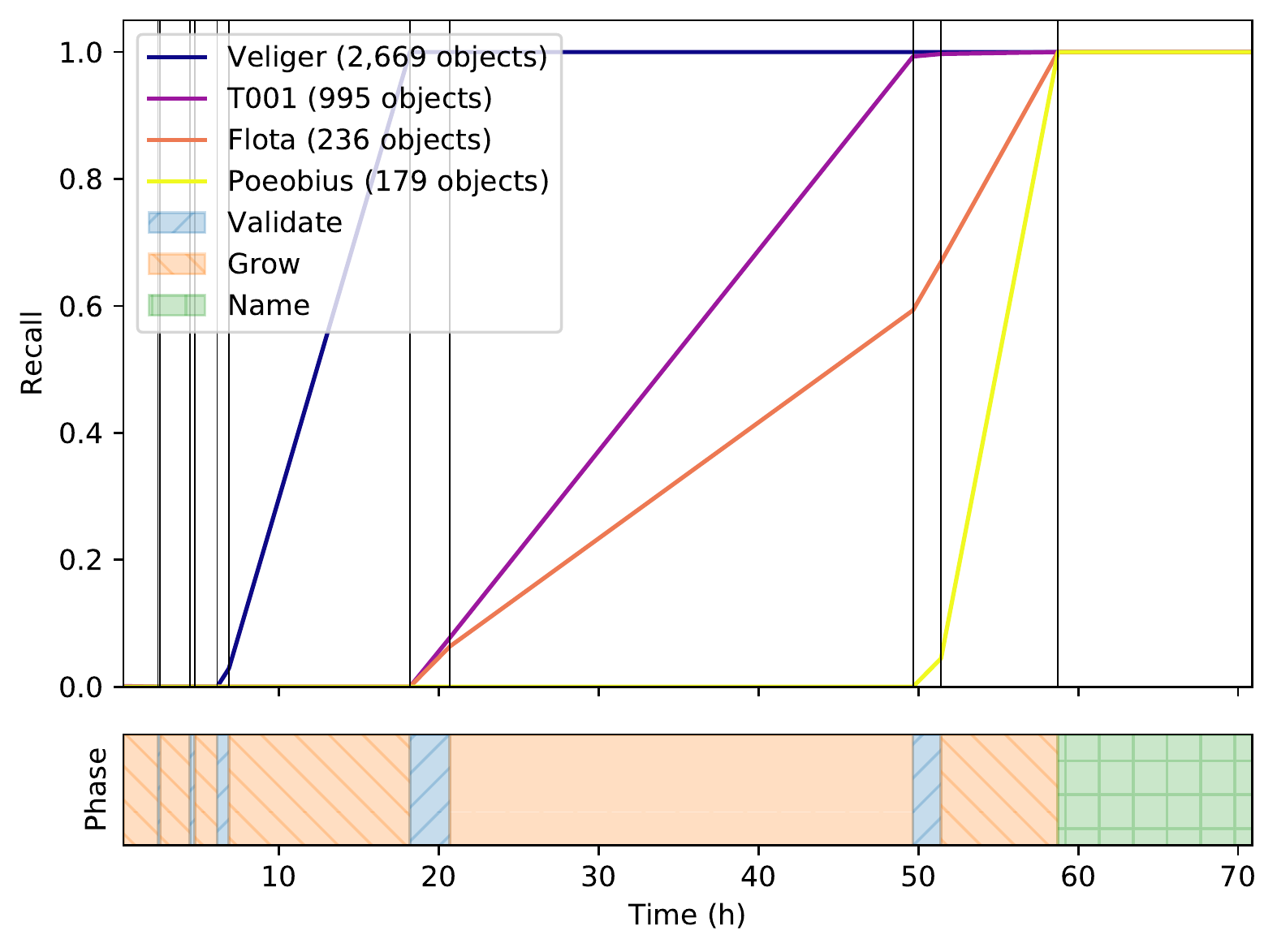}
		\captionof{figure}{Recall of the indicator classes \(\mathcal{C}_\text{i}\).
			The time periods are colored according to their respective phase.
			Veliger is found in the third iteration, T001 and Flota in the fourth,
			and Poeobius in the last iteration.}
		\label{fig:steps-phase-novel}
	\end{minipage}%
	\begin{minipage}[t]{.45\textwidth}
		\includegraphics[width=\textwidth]{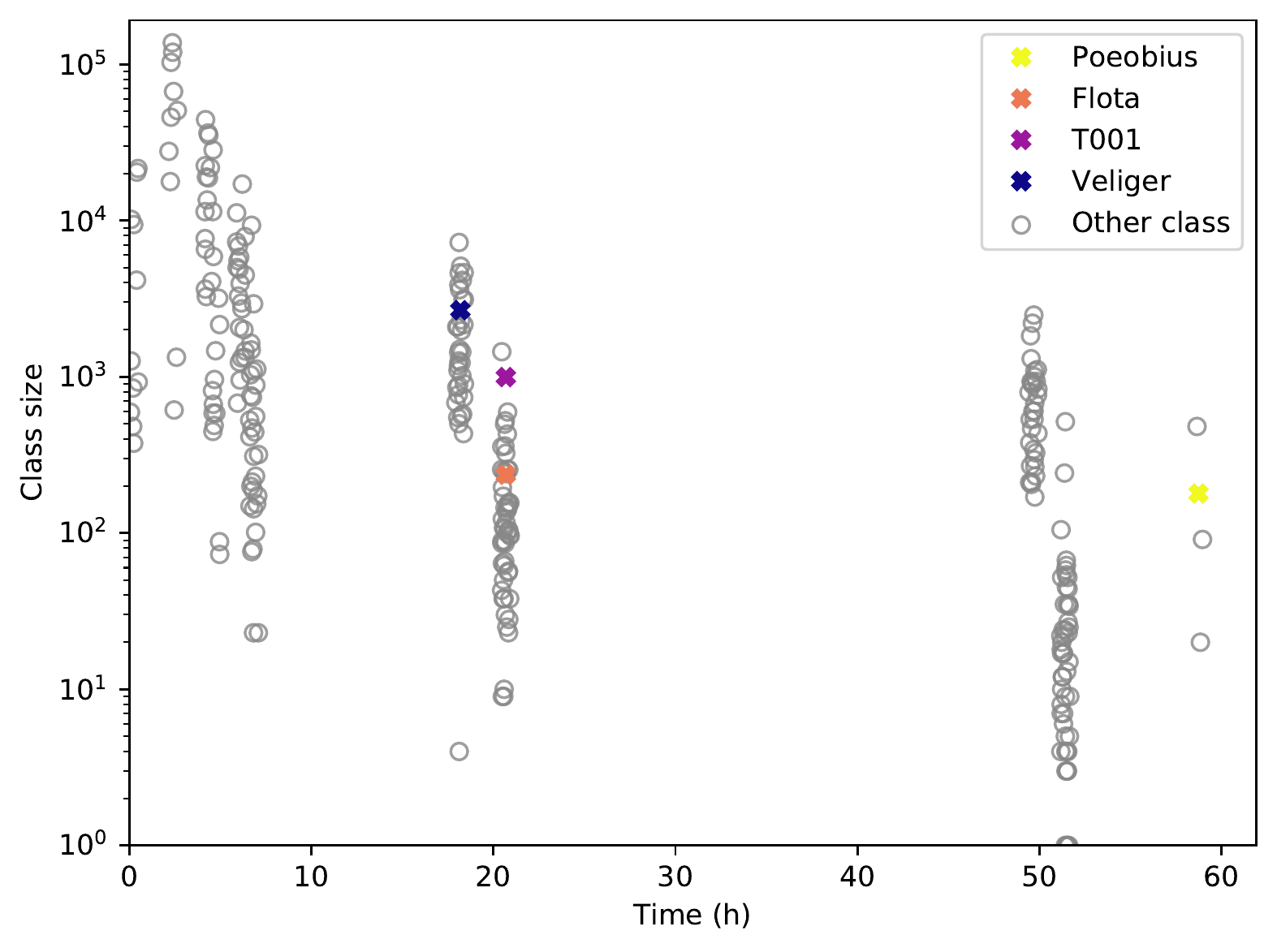}
		\captionof{figure}{Discovery of classes during the process.
		The larger a class, the earlier its seed (with at least \SI{5}{\percent} of the final number of objects) was found, as intended.}
		\label{fig:size-time}
	\end{minipage}
\end{figure}

The four held-out indicator classes \(\mathcal{C}_\text{i}\) were retrieved confidently,
meaning that they were the predominant class of at least one cluster, respectively.
\Cref{fig:steps-phase-novel} shows how Veliger, T001, Flota and Poeobius and the other classes started as very small cluster seeds and reached their final size throughout the processing of the data set.

\Cref{fig:size-time} illustrates the relationship between class size and time until retrieval:
As intended, larger classes were found in earlier iterations and the smaller a class, the later it was found during the process.
Veliger, the largest class with a very distinct shape, was retrieved early on.
Poeobius, the smallest of these four, was not found until the last iteration.
This trend is also reflected in the other classes.

\input{Discussion.tex}

\vspace{12pt}\noindent{\fontsize{9}{9}\selectfont\textbf{Data and code availability:}
{We release the source code for the MorphoCluster Application under the GNU General Public License (GPL) at \url{https://github.com/morphocluster}.
The initial training set \(\mathcal{L}_0\) and the new labeling \(\mathcal{L}_{MC}\) are available at \url{https://www.seanoe.org/data/00618/73002/}.}\par}

\authorcontributions{
	SM Schröder: Concept, Data curation, Formal analysis, Investigation, Methodology, Software, Visualization, Writing - original draft;
	R Kiko: Concept, Testing, Funding acquisition, Resources, Supervision, Writing - original draft;
	R Koch: Funding acquisition, Resources, Supervision;
}

\funding{%
This project was funded by the Cluster of Excellence 80 \enquote{Future Ocean} (CP1733).
\enquote{Future Ocean} is funded within the framework of the Excellence Initiative by the Deutsche Forschungsgemeinschaft (DFG) on behalf of the German federal and state governments. R.\ Kiko was furthermore supported by the SFB 754 \enquote{Climate-Biogeochemistry Interactions in the Tropical Ocean} (\url{www.sfb754.de},  grant no.\ 27542298 of the German Science Foundation DFG) and via a \enquote{Make Our Planet Great Again} grant of the French National Research Agency within the \enquote{Programme d'Investissements d'Avenir}; reference ANR-19-MPGA-0012.
}

\acknowledgments{We thank Svenja Christiansen,  Jean-Olivier Irisson and Marc Picheral for insightful discussions on plankton image sorting. Jean-Olivier Irisson and Marc Picheral furthermore provided background information about EcoTaxa.}

\conflictsofinterest{%
	The authors declare no conflict of interest.
	The funders had no role in the design of the study;
	in the collection, analyses, or interpretation of data;
	in the writing of the manuscript, or in the decision to publish the results.}

\externalbibliography{yes}
\bibliography{bib_cleaned}

\clearpage
\appendix

\section{Corresponding labels}%
\label{appendix:corresponding-labels}

This list accompanies \cref{fig:confusion-et-mc} and contains the corresponding names in
\(\mathcal{L}_0\) and \(\mathcal{L}_{MC}\).
The initial class names are in the same order as in the figure.
Manually established correspondences are printed in bold.

\input{correspondences.tex}

\end{document}

%% file: constants.tex
\def\constValGrowObjPH{20085.05986843072}
\def\constTotalDurationValGrowH{58.731714405000005}
\def\constTotalDurationNamingH{12.152545717777782}
\def\constTotalDurationNamingPercent{17.144209020068068}
\def\constNamingObjPH{97067.64552832354}
\def\constNamingClassesPH{23.0404399623358}

%% file: Introduction.tex
\section{Introduction}\label{introduction}

Current plankton imaging tools (e.g.~ZooScan~\cite{Gorsky2010}, UVP5~\cite{Picheral2010}, ISIIS~\cite{Cowen2008}, FlowCytoBot~\cite{Olson2007}, IFCB~\cite{Sosik2007})
deliver highly diverse and constantly growing plankton image data sets~\cite{Orenstein2015,Elineau2018}
that contain thousands, sometimes millions, of images sorted into a varying number of classes~\cite{Gonzalez2017a}.
It is expected that the volume and complexity of marine data will increase by orders of magnitude in the coming years~\cite{Malde2019a}.
Ecological analyses of these samples require accurate object counts to enable abundance estimates.
Object counts can be acquired by different means~\cite{Gonzalez2017}
but most often, each object is classified individually and the objects of each class are counted (\emph{classify-and-count}).
This confronts the field of marine ecology with the challenge of providing taxonomic identifications for enormous volumes of imaging data
efficiently. The annotation rate of human experts is long surpassed by the amount of data that is generated~\cite{Benfield2007}.
Therefore, advanced automatic image recognition techniques are indicated. These should liberate taxonomy experts from the tedious task of routine identifications~\cite{McLeod2010}.
However, to  extract valuable insights from the data, moderation of automatic techniques is imperative~\cite{Malde2019a}.

Published \emph{marine image annotation software}~\cite{Gomes-Pereira2016} tools include photoQuad~\cite{Trygonis2012}, VARS~\cite{Schlining2006}, Seascape~\cite{Teixido2011} and BIIGLE~\cite{Langenkamper2017}.
Beyond that, there are several tools not formally published like SQUIDLE+~\cite{squidle.org} and
EcoTaxa~\cite{Picheral2017} or the older Plankton Identifier~\cite{PlanktonIdentifier} and ZooImage~\cite{Bell2008}.
Some tools address the annotation of whole frames where objects of interest have to be localized first to be classified afterward.
However, plankton image data has usually a uniform background so no semantic segmentation is needed.
Other tools are therefore specifically targeted towards the annotation of individual Plankton images.

EcoTaxa~\cite{Picheral2017} is a web-application for the semi-automatic annotation of large image data sets of individual Plankton images.
We and other colleagues have been using it to sort UVP5 data for more than five years.
During this time, we noticed that we --- due to time constraints ---
often accept the automatic predictions for less interesting categories (\emph{default effect}),
we aggregate differently-looking objects due to taxonomic knowledge,
and focus only on the categories that are presumably relevant for the particular study.
For example, great effort went into the sorting of different Rhizaria~\cite{Biard2016} or finding instances of \textit{Poeobius}~sp.~\cite{Christiansen2018}.

Generally, researchers aim to annotate the objects according to a certain scientific goal, e.g.\ they sort all images of animals into accepted taxonomic units.
This means that e.g.\ different views (dorsal, lateral) of the same animal are grouped, although they might look very different.
Furthermore, taxonomic knowledge is applied when combining different taxonomic units into higher-order groupings (e.g.\ copepods, euphausiids and their larval stages into the subphylum crustacea).
On the other hand, a very detailed sorting of other parts of the data set is not done, although very different image classes do exist in this part.
Faecal pellets, aggregates and fibers might all be summarized under the term detritus.
Typically, only a few tens of classes are used in plankton studies based on imaging data~\cite{Schroder2019,Bochinski2019,Ellen2019,Orenstein2017,Ellen2015}
and the number of classes depends on the imaging instrument, sample location and research interest.
This \emph{interest-driven} data annotation approach --- that is also encouraged in EcoTaxa and other tools --- might be most feasible for exclusively manual annotation,
as it saves time, but it could be relatively problematic to automatically classify images into a set of so-defined classes.

Previously, shallow models like Support Vector Machines~\cite{Vapnik1998} or Random Forest~\cite{Breiman2001} with handcrafted local features measured on the image (e.g.\ size, gray level distribution, etc.) were used to classify plankton~\cite{Culverhouse1996,Blaschko2005,Sosik2007,Gorsky2010,Ellen2015}.
In recent years, however, there has been a transition towards deep plankton image recognition models based on convolutional neural networks (CNNs)~\cite{Lee2016,Orenstein2017,Graham2017,Christiansen2018,Ellen2019}.

Automatic classifiers require enough training data for each class.
Especially, all classes need to be known and well-represented in the training data.
Plankton image data contains a variety of dead matter, plankton of different size, morphology and orientation, and aggregations of multiple objects~\cite{Benfield2007} and is therefore a considerable challenge for image recognition.
This problem is further complicated because we observe a long-tailed abundance distribution of plankton in the wild~\cite{Schroder2019,Orenstein2017}.
While some of the ocean's inhabitants can be witnessed nearly everywhere, others are seldom seen at all.
Even if we knew which classes to expect in the sample, many could not possibly be represented in the training data because they were never annotated beforehand~\cite{Malde2019b}.
A classifier with a fixed set of classes prevents us from ever detecting anything new and unexpected. Such objects will be forced into the known classes and \enquote{disappear}.
Therefore, the analysis can only provide insights that are compatible with the initial question and classification granularity
and does not necessarily extend to the full information which the current sample actually provides.

Apart from them not being complete, reliance on training sets has further weaknesses:
First, they might deviate from the distribution of the collected sample.
In the case of \emph{classify-and-count}, this could in some cases distort the abundance estimates severely~\cite{Gonzalez2017}.
Second, a consensus on the identification of objects is hard to obtain in practice~\cite{Culverhouse2007},
so training sets --- like every collection of annotated real-world data --- exhibit some inconsistencies.

Consequently, the incoming data has to be constantly monitored,
meaning that the automatic classifications are often manually validated by experts~\cite{Picheral2017}.
Given the growing amount of data, this will prove less and less feasible.
In \cite{Christiansen2018}, the polychaete \textit{Poeobius}~sp. was only found in an Underwater Vision Profile 5 data set, after it was seen in underwater videos taken in parallel with the PELAGIOS~\cite{Hoving2019}.
A mostly manual examination of 1.8M UVP5 images from the Eastern Tropical Atlantic then yielded 450 images of \textit{Poeobius}~sp.

When objects are sorted manually, several human factors like cognitive biases, fatigue and boredom~\cite{Culverhouse2007} influence the classification.

These factors altogether --- dependence on training data, a fixed set of classes, changing long-tailed distributions, growing amounts of data, and adversarial human factors --- limit the accuracy and utility of interest-driven data annotation.
Instead, we argue for \emph{data-driven} image sorting using \emph{unsupervised machine learning techniques} in order to be able to define all classes in the data set, to spot novelties and unexpected patterns and derive reliable abundance estimates.

\subsection{MorphoCluster}

In this work, we present \emph{MorphoCluster}, a tool for data-driven, fast and accurate annotation of large data sets of single object images.
Although we present and discuss the tool in the context of marine image annotation, it should be applicable in many areas with similar data sets (images of individual objects).

Considering the strength of deep neural networks to learn distinctive features~\cite{Oquab2014},
we hypothesize that it is feasible to cluster these features to partition a plankton image data set in a meaningful way.

We therefore combine unsupervised clustering with an interactive tool to revise the initial clusters, arrange them hierarchically,
manually correct the hierarchy and annotate the clusters. 
The annotator therefore can explore the groupings inherent in the data and spot novelties and unexpected patterns.
By annotating groups of similar images as a whole, we intend to enable the consistent manual review of large amounts of data in a rather short time.

In the following, we will show that by paying attention to the cluster structure of a data set, MorphoCluster is at the same time fast, accurate and consistent,
provides a fine-grained and data-driven classification and enables novelty detection.

%% file: project_phases_simple.tex
\begin{tabular}{cc @{\hspace{4\tabcolsep}} cS[round-mode=places, round-precision=2,table-format=1.2]S[round-mode=places, round-precision=0,table-format=6.0]}
\toprule
{Iteration} & {$m$} & {New clusters} & {Validated clusters} & {Objects sorted per hour} \\
\midrule
          1 &   128 &             37 &                   26 &             195778.619994 \\
          2 &    64 &             51 &                   49 &             144558.700934 \\
          3 &    32 &            110 &                  100 &              93332.801008 \\
          4 &    16 &            299 &                  288 &              14875.430429 \\
          5 &     8 &            447 &                  438 &               2282.304830 \\
          6 &     4 &            612 &                  291 &                834.422336 \\
\midrule
\multicolumn{2}{c}{Total} & 1556 & 1192 & 20085.05986843072 \\
\bottomrule
\end{tabular}

%% file: empty_correspondences.tex
\emph{Annelida\_Polychaeta}, \emph{Crustacea\_leg}, \emph{Diplostraca\_Cladocera}, \emph{Euopisthobranchia\_Thecosomata}, \emph{Mollusca\_Cephalopoda}, \emph{Pyrosomatida\_Pyrosoma}, \emph{Solmundella\_Solmundella bitentaculata}, \emph{detritus\_light}, \emph{othertocheck\_darksphere}, \emph{temporary\_t009}

%% file: Discussion.tex
\section{Discussion}%
\label{sec:discussion}

Imaging applications spread as prices for camera systems decline and technological
advancements allow for autonomous deployments. Within plankton research --- but also in many other
domains --- we face a flood of image data that requires interpretation~\cite{Lombard2019}.
While supervised machine learning approaches are generally very fast and can be very accurate,
they are limited to a fixed classification scheme, so without further measures, they fail at novelty detection~\cite{Shu2018},
and might perpetuate biases from the training set~\cite{Kotsiantis2006}.
Humans on the other hand excel at fine-grained object classification and novelty
detection but are limited in their annotation rate.
Their speed and accuracy are impaired by fatigue or boredom and cognitive biases, as they might favor a recently used label (recency effect)~\cite{Culverhouse2007} or an automatic prediction (default effect).
Thus we need to develop techniques that exploit and augment the human ability to perform object classification and novelty detection by accelerating annotation and increasing consistency~\cite{McLeod2010}.

MorphoCluster excels at cluster-based manual mass allocation of images into homogeneous
groups, followed by hierarchical ordering in a semantic tree for easy naming of classes.
By paying attention to the cluster structure of a data set,
we achieve an outstanding combination of properties:
MorphoCluster is at the same time fast, allows for a flexible, fine-grained and data-driven classification, is accurate, consistent,
and enables novelty detection. MorphoCluster is available as open-source software at \url{https://github.com/morphocluster}.
We expect that the approach can be adapted to any kind of image collection where individual objects can be extracted and useful features that enable meaningful clustering can be calculated using a deep convolutional neural network (CNN).

\subsection{Feature extraction and clustering using deep learning approaches}

CNNs can generate features that are powerful and general enough to perform classification tasks using shallow classifiers like random forests, support vector machines, or logistic regression~\cite{Orenstein2017,Picheral2017,VanGinneken2015,Razavian2014,Donahue2013},
consistently outperforming hand-crafted features~\cite{Razavian2014}.
Malde and Kim~\cite{Malde2019b} show --- by using some selected categories from a well-sorted data set --- that features extracted with a siamese network can also be used to cluster images into relevant categories and allow for nearest neighbor and closest centroid classification.
CNN image features also enable clustering into semantic categories on which the network was never explicitly trained~\cite{Guerin2018,Donahue2013}.
Features learned on one task (e.g. natural objects like birds, horses and sheep) are also often transferable to a different task (e.g. the distinction of man-made objects like bicycles, cars and trains)~\cite{Orenstein2017,Simonyan2015,Yosinski2014}.
We therefore tested in some preliminary experiments if we could train a feature extractor with ImageNet~\cite{Deng2009} data.
However, this did not produce well-defined clusters and we fine-tuned the network with plankton images so it could learn the characteristic appearances of different kinds of plankton.
The CNN features extracted using this auxiliary training set then allowed efficient clustering and transformation into a hierarchy by agglomerative clustering.
 
We use the advantageous characteristics of the CNN features to provide a complete workflow to separate and classify plankton images in a real-world data set.
By merging supervised and unsupervised tools with human intervention, MorphoCluster enables flexible, fine-grained mass annotation of images and detection of novel classes in a data-driven way.

\subsection{MorphoCluster is data-driven}%
\label{subsec:data-driven}

Image classification is often \emph{interest-driven}, i.e.\ driven by prior knowledge and expectations of the data, which is reflected in the routinely small number of classes used~\cite{Bochinski2019,Ellen2015}.
The applied classification scheme is then based on a certain research question and the annotation effort is largely influenced by this question as well.
Accordingly, some \enquote{interesting} object types are sorted with high effort,
some \enquote{less interesting} types are subsumed in general classes.
Furthermore, classification methods typically assume that training data and test data are independent and identically distributed~\cite{Gonzalez2017,Forman2008}.
However, this is often not the case as distribution patterns change with temporal (e.g. seasonal) and spatial dynamics~\cite{Christiansen2018,Mackas2012} and can therefore be different for each sample~\cite{Gonzalez2017,Gonzalez2017a}.
Because classifiers are optimized for the distribution of the training sample and inherit their biases, their prediction might not represent the true data distribution of a test sample~\cite{Gonzalez2017a}.

Computer-aided image classification tools (e.g.\ EcoTaxa~\cite{Picheral2017}, SQUIDLE+~\cite{squidle.org}, Pl@ntNet-Identify~\cite{Joly2014} and others~\cite{Waldchen2018a,Waldchen2018}) assume that most images can be sorted into a set of classes that are defined beforehand or \emph{ad hoc}.
Furthermore, predictions might be skewed towards the class proportions of the training set and objects are predicted into a similar but incorrect category.
Annotators might then tend to accept the prediction when they feel no strong preference (\emph{default effect}).
On the other hand, because of the \emph{contrast effect}, an annotator might move objects, that are correctly predicted as one class
(e.g.\ \enquote{detritus\_dark}) but are in some property different (e.g.\ lighter) than the other displayed objects surrounding them,
to another (incorrect) class (e.g.\ \enquote{detritus\_light}).
Interest-driven sorting using conventional tools is therefore sometimes rather subjective and might cause a certain blindness towards
the nuances in the data. 

While an annotator working with MorphoCluster is still influenced by the same cognitive biases,
these biases have different effects than during the usage of conventional tools.
MorphoCluster allows sorting data without a preconception about the relative class abundance and takes a \emph{data-driven}, explorative, yet manually controlled image annotation approach.
Creating classes from homogeneous clusters in our view fits the granularity of the data set itself well.
This approach minimizes negative subjective influences and makes structures in the data visible.
The impact of the default effect is less pronounced:
During cluster validation, an annotator might be tempted to just accept the proposed cluster which would impair sorting accuracy if the cluster is not clean.
Due to the simplicity of the task (homogeneous / not homogeneous), however, the problem should not be as severe as with conventional sorting.
The contrast effect is actually exploited to reject clusters with major impurities by showing dissimilar images side by side.
In case that a meaningful cluster is rejected (e.g. in the second round of clustering and growing), this will slow down the process but will not affect the final result. This cluster should be proposed again in the subsequent round of clustering and growing and will still be detected. Therefore, the annotator is bothered by little remorse to reject a cluster during cluster approval.
Also during growing, we use the contrast effect to our benefit as we oppose the cluster seeds and the images to be added to the cluster.
Strong differences therefore can be easily spotted.
We introduced the \enquote{turtle mode} to make the acceptance or rejection of images at the cluster borders more flexible.
Especially bulk acceptance might be a problem due to the default effect, whereas bulk rejection will only slow down the process.
Contrast, default and recency effect should have little impact during cluster annotation in the hierarchic arrangement of the last step of MorphoCluster.
The hierarchic arrangement is data-driven and we observe that similar clusters are located in according branches.
An annotator might keep branches of the automatic hierarchy (default and recency effect) until a strong contrast is found.
Nuances in the data set therefore might be overlooked, but as only comparatively few clusters need to be named, the decisions are few and can be made with great care.
In general, fatigue and boredom during cluster approval, growth and naming is in our view much reduced in comparison to conventional sorting.
The cognitive demanding classification task to allocate a name to a given object needs to be executed only in comparatively few cases, whereas the detection of new or exceptionally large clusters can be perceived as especially rewarding.
As with any sorting tool, appropriateness of the sorting and annotation in MorphoCluster finally depends on the care the annotator assigns to the task.
We nevertheless expect the results to be rather objective as the annotator is guided by the data structure and mostly needs to execute simple and effective tasks.

\subsection{MorphoCluster is fast}

Our strategy transforms time-consuming image annotation of single images into the much faster annotation of clusters.

For manual or prediction-based tools, sorting time depends on the number of objects and the number of classes~\cite{Tian2007},
but details on effort and speed required to sort a data set are often not reported in the literature (e.g.\ \cite{Bochinski2019,Orenstein2015,Gorsky2010,Deng2009}).
With overall nearly 17k objects per hour, MorphoCluster reaches or even surpasses the sorting speed of the well-optimized
supervised classification approach implemented in EcoTaxa (\cite{Picheral2017}, pers.\ comm.).
Depending on the size and complexity of a project,
EcoTaxa allows sorting speeds between approximately 300 and 15k objects per hour.
Typically, objects are automatically classified in EcoTaxa, then the predicted images for each class are manually validated.
The validation of predictions with high classification scores is commonly fast while low classification scores require extensive manual resorting. In the first iterations of the MorphoCluster process, the sorting speed can reach 200k objects per hour, whereas it also slows down when cluster sizes decline. Most projects in EcoTaxa use up to 90 annotation categories (pers.\ comm.), substantially less than those that emerged in MorphoCluster. It is known that it takes longer to pick a category from a larger menu~\cite{Fasolo2009}, which indicates that the difference in sorting speed between EcoTaxa and MorphoCluster might be larger if the same granularity would be targeted.

The authors of \cite{Tian2007} propose a face annotation framework that, like MorphoCluster,
uses partial clustering and subsequent annotation of clusters and remaining data to quickly label large amounts of face images.
In agreement with our results, they observe that clustering can substantially reduce the annotation workload
because each user interaction affects a large number of individual objects
and partial clustering groups images into meaningful and homogeneous clusters.
They provide a rough estimate that their approach is 5 times as fast as conventional sorting.

To increase the overall speed of MorphoCluster, we optimized each individual step.
During validation, clusters of similar objects are accepted as a whole which drastically reduces the number of entities that require annotation in further steps.
In the cluster growth step, binary search enables the user to quickly find the border of a cluster.
Thus, adding any number of objects to a cluster requires only a small fraction of the time required to annotate these objects individually.
When the border of the cluster is reached, the user can also delete or accept single images, which activates a \enquote{turtle mode}, disables binary search and forces the user to conduct single image approval.
The suitability of our cluster growth strategy is clearly confirmed by the high sorting accuracy.
We investigated if the growth of the clusters could be optimized by accounting for non-spherical clusters, but noticed no improvement.
The hierarchical arrangement of similar clusters facilitates their naming.
The same time to identify a single object in traditional approaches is spent to identify many objects, sometimes even thousands,
which in turn leads to less time pressure in assigning proper names.
MorphoCluster's high sorting rate is a result of the fact that simple user decisions in each step affect a large
number of objects and as partitioning and naming are different steps, more effort can be put into a precise and fine-grained classification.

\subsection{MorphoCluster provides a flexible and fine-grained classification}

For MorphoCluster, we developed a strategy for cluster retrieval that guarantees that large clusters are retrieved at the beginning of the process and small clusters only at the end.
Preliminary experiments showed that settings that allow for small cluster sizes immediately lead to an over-separation of some classes and fragmented larger classes into many more or less indistinguishable clusters.
These mostly consisted of some detritus categories.
Merging and/or naming of these clusters would have become very time consuming and in very many cases we would have given identical names for these clusters.
Our strategy to first retrieve large clusters improved the situation, but still, some clusters were retrieved that were subsequently merged during the naming step.
Our hierarchical naming tool nevertheless makes these decisions less subjective, as it contrasts similar clusters.
In the end, the decision of whether or not two groups of images show the same category is made by the user.
Further research is necessary to optimize the strategy of cluster retrieval and growth as an optimal path through the data should exist that could reduce the need to merge clusters.
In comparison to the original data set which was sorted into \num{65} classes, we retrieved \num{280} classes and in general a more fine-grained sorting, which might reveal new insights.
Detritus, for example, was previously often sorted into less than ten classes, although there can be strong differences in shape and size which are likely related to its biogeochemical properties.
A nuanced isolation of these shapes makes it easier to find such properties in data.

\subsection{MorphoCluster enables detection of novel classes}

As data sets increase in size, former outliers may grow into new categories:
Consider a data set containing 1k images.
It might contain a single image of \textit{Poeobius}~sp., a species found in very low numbers throughout the whole Atlantic Ocean which under certain conditions proliferates strongly~\cite{Christiansen2018}.
Sorting the whole data set by hand, an expert would create a class \enquote{Poeobius} because of their knowledge of its appearance.
Another possibility is that these images are subsumed under a more general category during interest-driven sorting.
Using MorphoCluster, we would not find this single image, because MorphoCluster is geared towards finding groups of similar objects.
If we now collect more images from the same source and grow this image data set,
the number of \textit{Poeobius}~sp. images might grow proportionally and we should find 1000 images in a 1 Million image data set.
Our experiment indicates that these images would then be found as a cluster that can be identified and named.

MorphoCluster's data-driven approach allowed the reliable detection of the held-out indicator classes (Veliger, T001, Flota and Poeobius) and we predict that by applying the natural decision boundaries dictated by the density
structure of the data it is equally likely to find other novel classes.
Several of the transitional classes we identified (like depicted in \cref{fig:transitional}) could also be considered novel classes.

Therefore, we deem MorphoCluster well-suited to search the numerous sources of constantly growing marine imaging data for previously undocumented categories.

\subsection{MorphoCluster is accurate and consistent}

The accuracy of human sorting mainly depends on the operator.
Within plankton research, experts can reach a panel consistency of up to \SI{95}{\percent} for small numbers of categories~\cite{Culverhouse2003}.
Using MorphoCluster, most of the resulting \num{280} classes were sorted with very high consistency in the same range (see \cref{subsec:accuracy})
and similar-looking objects share the same annotation.
This can be explained by the fact that the MorphoCluster process starts with very homogeneous clusters of objects that stay homogeneous even after growing.
As discussed previously, a user is less affected by cognitive biases when using MorphoCluster than when using conventional methods.
This way, the homogeneity of clusters is carried through to the end of the whole process.

In manual or prediction-based sorting tools, objects are typically sorted individually and the context of
similar objects is not available.
Conversely, clustering-based approaches provide this kind of context by constructing homogeneous groups of objects~\cite{Tian2007},
a huge advantage that is also shared by MorphoCluster.

\subsection{Possible improvements of MorphoCluster}

\subsubsection{Feature learning and clustering}

Feature learning and clustering are sequential steps in the current MorphoCluster process and we rely on an initial training set to train the feature extractor.
Recent works on unsupervised learning of deep image descriptors combine feature learning and clustering and do not require any labels~\cite{Aljalbout2018,Haeusser2018,Caron2018,Xie2016a,Yang2016}.
These unsupervised feature learning methods could be investigated to reduce the reliance on labeled data.

A small number of objects was ultimately left untreated (residual objects) and a handful of known small classes was not retrieved.
An adjustment of the feature extractor or the use of a different clustering algorithm could maybe help to mitigate this problem.
Still, it is obvious that classes with a very small number of objects (\emph{low-shot} or \emph{one-shot classes}~\cite{Vinyals2016,Finn2017}) can not be retrieved by clustering although human knowledge indicates their presence.
To facilitate their retrieval, \emph{spiking} the unlabeled data with labeled objects could increase their density in the feature space
and low-shot learning techniques\cite{Schroder2019} could be employed to identify them prior to clustering but this does not work for unknown classes.
Therefore, methods of novelty detection~\cite{Pimentel2014} (e.g.\ \cite{Bodesheim2015}) should be investigated.

One of the classes not retrieved using MorphoCluster, \textit{Pyrosoma}~sp.\ (named  \emph{Pyrosomatida\_Pyrosoma}), exhibits some very large images.
Large variations in image size are a general problem for convolutional neural networks.
To be able to process these images, we scale the images down to the input size of the network.
Unfortunately, this can weaken and sometimes even remove their distinctive features.
A possible future research direction is therefore the exploration of
attention mechanisms~\cite{Sun2020,Sun2019,Zheng2019} that allow the network to focus on specific image regions
and view them in full resolution.
Some distinguishing features of an object might not be represented in the features learned by the deep feature extractor,
either because of insufficient sensor resolution or because they are of a different modality (e.g.\ genetic, environmental, \ldots).
The introduction of other morphometric~\cite{Campbell2020} and environmental~\cite{Ellen2019} information into the deep learning image recognition could therefore be a viable option to improve clustering and reduce the number of residual objects.

The HDBSCAN* algorithm that was used in this work has a runtime super-linear in the number of objects and the number of dimensions at best~\cite{McInnes2017}.
Speeding up the clustering approach could enable the execution of the clustering, growing and approval procedure in single rounds
so that only the largest and best-defined cluster is extracted in every iteration and thereby enable a more interactive user experience.
This would especially be useful at the beginning of the procedure as it would yield a more optimal path through the data.
The main competitor is k-means with a best-case runtime linear in the number of objects and the number of dimensions~\cite{McInnes2017}, which becomes quite an advantage with large data volumes.
However, k-means is a \emph{partitioning} clustering algorithm, while HDBSCAN* does not necessarily assign a cluster for all points, and the question remains how it can be adapted to the requirements of the MorphoCluster framework.

\subsubsection{Hierarchical naming}

Although the morphology of an organism is in part determined by its genes, this relationship is very complex.
As an example, larvae and adults can look completely different although they share the same set of genes~\cite{Kiko2008}.
The class hierarchy that we used as a starting point in the naming step was generated from the list of clusters using agglomerative clustering which successively contracts similar clusters~\cite[p.~73]{Everitt2011}.

The calculated cluster hierarchy coincides only in few cases with the known phylogenetic tree of life
because the phylogenetic tree is derived not only from images but also, for example, from genetic, ontogenetic and microscopic analysis.
We chose average linkage (UPGMA) clustering as a robust default method and it should be investigated if alternatives (e.g.\ WPGMA~\cite[p.~79]{Everitt2011}) lead to a closer match between precomputed hierarchy and manually tuned end result.

The final sorting emerges from the interaction of the taxonomic knowledge of the annotator and the data-driven arrangement of the data set.
This interaction could be further facilitated by including an extensible reference taxonomy in the application, spiking the input data with existing labeled data to match the emerging clusters to known classes (like we did in the evaluation of our approach), or providing some sort of vocabulary to avoid the occasional naming inconsistencies introduced by the free-form input.
It also seems useful to use the clusters from a first MorphoCluster run as seeds in future runs, which only need to be grown using the new data.

\subsubsection{Division of labor}
MorphoCluster could enable a unique distribution of efforts between users with different expertise to accelerate sorting and make better use of available human resources.
The separation of sorting and naming could allow entrusting the relatively simple task of validating and growing homogeneous clusters to less experienced staff, while professional taxonomists, whose time is a precious resource~\cite{McLeod2010}, could focus on the more complex but less time-consuming task of cluster identification.

Multi-user approaches during which several users work on different clusters of a given data set should also be possible. The high throughput of MorphoCluster could even enable the replication of the entire process by different experts or teams, which should increase the overall annotation quality even further.

\subsection{Conclusions}

With MorphoCluster we present a novel approach to image annotation that does not require the user to take a look at \emph{every single image}.
Rather, similar images are automatically aggregated in clusters, which are checked for consistency.
These clusters are thereafter grown and named \emph{de novo}, avoiding biases of a given prediction or sorting scheme.
We succeeded to shift the unit of labor during the sorting process from individual images to often very large clusters.
The development of useful CNN features was in our view critical for this success.
The result of our efforts is a simple and fast manual annotation tool, which yields a consistent and fine-grained sorting.
The sorting effort with MorphoCluster scales primarily with the number of classes of a given data set while with other tools the effort scales with the number of images.
We argue that our approach is less biased by contrast, default and recency effects and avoids pitfalls of interest-driven sorting.
The primary use case for MorphoCluster is the \emph{rapid annotation of images} to acquire huge volumes of labeled data for further data analysis or to initialize a training set.
Importantly, MorphoCluster also enables novelty detection and facilitates the data-driven creation of possibly meaningful subcategories.
By using MorphoCluster, we can shift away from accidental discoveries and a lot of manual labor to a systematic and fast strategy for surveying the ocean.
It will hopefully help to stem the flood of plankton image data that we expect and may be just as useful for annotating other image data sets.

%% file: correspondences.tex
\begin{description}
  \footnotesize
  \item[Annelida\_Polychaeta (0):] {\it n/a}
  \item[Crustacea\_leg (0):] {\it n/a}
  \item[Diplostraca\_Cladocera (0):] {\it n/a}
  \item[Euopisthobranchia\_Thecosomata (0):] {\it n/a}
  \item[Mollusca\_Cephalopoda (0):] {\it n/a}
  \item[Pyrosomatida\_Pyrosoma (0):] {\it n/a}
  \item[Solmundella\_Solmundella bitentaculata (0):] {\it n/a}
  \item[detritus\_light (0):] {\it n/a}
  \item[othertocheck\_darksphere (0):] {\it n/a}
  \item[temporary\_t009 (0):] {\it n/a}
  \item[Appendicularia\_body (1):] {aggregate/aggregate-balls-grey\_to\_aggregate-fluffy-grey}
  \item[Arthropoda\_Crustacea (1):] {crustacea/spider-like-amphipods}
  \item[Collodaria\_solitaryfuzzy (1):] {rhizaria/solitary-black-like\_to\_rhizaria}
  \item[Euopisthobranchia\_Gymnosomata (1):] {fiber/fiber-boundles-grey\_to\_aggregates-fluffy-grey}
  \item[Munididae\_Pleuroncodes (1):] {\bf crustacea/pleuroncodes}
  \item[Terebellida\_Flota (1):] {\bf polychaeta/flota}
  \item[Thaliacea\_Salpida (1):] {cut/cut-aggregates-jellies}
  \item[Trachylina\_Narcomedusae (1):] {cnidaria/medusa-large\_to\_cut}
  \item[temporary\_t002 (1):] {\bf rhizaria/eight-armed}
  \item[temporary\_t003 (1):] {\bf rhizaria/six-lobes}
  \item[temporary\_t005 (1):] {\bf rhizaria/sphere-thorns-w-balls}
  \item[temporary\_t006 (1):] {\bf metazoa/salpida-larvae}
  \item[temporary\_t015 (1):] {aggregate/aggregate-fluffy-dark\_to\_fiber-fluffy-dark}
  \item[Appendicularia\_house (2):] {aggregate/aggregate-large-fluffy-and-appendicularia}, {aggregate/very-fluffy-possibly-discarded-appendicularia-houses}
  \item[Hydroidolina\_Siphonophorae (2):] {artefact/cut}, {aggregate/aggregate-very-large-w-cut-tentacles}
  \item[Metazoa\_Ctenophora (2):] {cut/cut-aggregates-turbid}, {ctenophora/beroe-type}
  \item[Vertebrata\_Gnathostomata (2):] {\bf metazoa/fish}, {chaetognatha/chaetognatha\_to\_cut}
  \item[temporary\_t004 (2):] {\bf rhizaria/triangular-sphere}, {rhizaria/triangular-eye-w-spikes}
  \item[temporary\_t010 (2):] {aggregate/aggregate\_dark\_thorny\_to\_fiber}, {feces/feces-little-fluffy}
  \item[temporary\_t012 (2):] {aggregate/aggregate-fluffy-lightgrey}, {feces/feces-straight-faint-fluffy}
  \item[Aulacanthidae\_Aulacantha (3):] {\bf rhizaria/sphere-thorn}, {rhizaria/rhizaria-mix}, {rhizaria/sphere-thorn\_to\_legs}
  \item[Metazoa\_Mollusca (3):] {compact/compact-dark\_to\_aggregate-dark-fluffy}, {crustacea/crustacea-like}, {rhizaria/double-lobes}
  \item[Phaeosphaerida\_Aulosphaeridae (3):] {\bf rhizaria/sphere-eye-w-spikes}, {solitary-black/solitary-black-small-w-grey-surrounds}, {sphere-thorn/sphere-thorn\_to\_badfocus}
  \item[Terebellida\_Poeobius (3):] {\bf polychaeta/poeobius}, {compact/compact-doubles-w-fluffy-surrounds}, {fiber/fiber-bundle}
  \item[Tunicata\_Appendicularia (3):] {aggregate/aggregate-large-marine-snow-w-black-parts\_to\_cut}, {aggregate/aggregate-very-fluffy-and-large}, {appendicularia/appendicularia-hous}
  \item[artefact\_turbid (3):] {\bf artefact/turbid}, {turbid/turbid-fuzzy}, {turbid/turbid-w-objects}
  \item[detritus\_compact (3):] {aggregate/aggregate-fluff-dark-edges}, {compact/compact-dark}, {compact/compact-small-dark\_to\_crustacea}
  \item[temporary\_t001 (3):] {fiber/fiber-loops}, {rhizaria/almond-eye}, {unknown/halfmoon-w-dot}
  \item[Hydrozoa\_Cnidaria (4):] {aggregate/aggregate-very-fluffy}, {cnidaria/fringed-jellies-w-dot}, {cnidaria/jellies-w-cross}, {cnidaria/jellies-w-stripes}
  \item[Retaria\_Foraminifera (4):] {cut/cut-tentacles}, {foraminifera/foraminifera-tight}, {foraminifera/foraminifera\_to\_foraminifera-cut}, {rhizaria/foraminifera\_to\_spiky}
  \item[Metazoa\_Chaetognatha (5):] {\bf metazoa/chaetognatha}, {fiber/fiber-fluffy-grey}, {fiber/fiber-thin\_to\_chaetognatha}, {chaetognatha/chaetognatha\_to\_badfocus}, {mix/fiber\_tentacles}
  \item[Mollusca\_veliger (5):] {\bf mollusca/veliger}, {aggregate/aggregate\_to\_solitary-black}, {aggregate/small-aggregates-mixed-w-compact}, {detritus/rhizaria-remains\_to\_aggregate\_small\_fluffy}, {veliger/veliger-straight-arms}
  \item[Oligostraca\_Ostracoda (5):] {\bf crustacea/ostracoda}, {aggregate/aggregate-angled-grey}, {compact/compact-dark-thorny}, {copepoda/copepoda-side-view\_to\_detritus-compact-grey}, {mix/mix-of-different-grey-items}
  \item[Retaria\_Acantharea (5):] {\bf rhizaria/acantharia}, {aggregate/aggregate-compact-large-grey}, {fiber/fiber\_to\_solitary-black}, {rhizaria/acantharia\_to\_spiky}, {tuft/tuft-grey-irregular}
  \item[detritus\_ovoid (5):] {aggregate/aggregate-compact-grey}, {aggregate/aggregate-rings-small-dark}, {compact/compact-grey-egg-form}, {compact/compact-grey-slightly-fluffy}, {unknown/eye-slit}
  \item[fiber\_fluffy (5):] {aggregate/aggregate-compact-fluffy\_to\_feces}, {aggregate/aggregate-fluffy-w-two-dots}, {aggregate/aggregate-very-loose-w-dark-spots}, {fiber/fiber-fluffy-w-dark-spots}, {unkown/appendicularia\_body\_to\_harpacticoid}
  \item[fluffy\_light (5):] {aggregate/aggregate-fluffy-loose-grey\_to\_fluffy\_fiber}, {aggregate/aggregate-fluffy-loose-lightgrey}, {aggregate/aggregate-very-fluffy-loose-grey}, {mix/detritus\_to\_crustacea}, {unknown/ovoid-w-dot}
  \item[temporary\_t011 (5):] {aggregate/aggregate-fiber-fluffy-dark}, {aggregate/aggregate-fiber-fluffy-grey}, {aggregate/aggregate-fluffy-dark}, {aggregate/aggregate-fluffy-lightgrey\_to\_feces}, {aggregate/aggregate-thorny-grey\_to\_crustacea}
  \item[Collodaria\_collonial (6):] {\bf collodaria/colonial}, {aggregate/aggregate-fluffy-large-compact-maybe-from-appendicularia}, {aggregate/aggregate-large-ball\_to\_globule}, {aggregate/aggregate-very-loose-grey}, {sphere-thorn/sphere-thorn-doubles}, {unkown/small-rings-w-dots}
  \item[Collodaria\_solitaryblack (6):] {\bf rhizaria/solitary-black}, {compact/compact-round-black\_to\_solitary-black}, {rhizaria/rhizaria\_to\_detritus}, {rhizaria/solitary-black-like\_to\_acantharia}, {solitary-black/solitary-black-large}, {solitary-black/solitary-black\_to\_sphere\_eye}
  \item[Collodaria\_solitarygrey (6):] {aggregate/aggregate-ball-dark}, {aggregate/aggregate-fluffy-w-many-dark-spots}, {compact/compact-small-round-grey\_to\_solitary-black}, {acantharia/acantharia-small\_to\_solitary-black}, {rhizaria/rhizaria-small\_to\_compact}, {rhizaria/solitary-black-faint}
  \item[Metazoa\_Cnidaria (6):] {cut/cut-jellies}, {aggregate/aggregate-balls-fluffy-grey}, {cnidaria/bitentaculata}, {cnidaria/jellies-large-medusa}, {cnidaria/jellies-w-dot-and-edges}, {cnidaria/round-jellies-w-dot}
  \item[Trichodesmium\_tuff (6):] {feces/feces-short-grey\_to\_trichodesmium-tuft}, {fiber/fiber-bundle-grey-small}, {puff/puff-large\_to\_fiber-bundle}, {tuft/feathery-ending}, {tuft/feathery-sharp-ending}, {tuft/sharp-ending}
  \item[Cnidaria\_Hydrozoa (7):] {aggregate/aggregate-compact-fluffy-grey}, {aggregate/aggregate-fluffy-light}, {compact/compact-grey-w-small-jellies\_to\_aggregate}, {fiber/fiber-bended-ring-like}, {cnidaria/jelly-small-rings\_to\_badcocus}, {cnidaria/small-jellies-w-dot}, {unknown}
  \item[Phaeodaria\_leg (7):] {\bf rhizaria/foraminifera\_to\_sphere-legs}, {detritus/acantharia-remains\_to\_fiber}, {fiber/fiber-bundle-small}, {fiber/fiber\_to\_rhizaria-spiky}, {rhizaria/foraminifera\_to\_fiber-bundle-fluffy}, {rhizaria/solitary-black-like}, {rhizaria/triangular-sphere\_to\_sphere-legs}
  \item[Collodaria\_solitaryglobule (8):] {aggregate/aggregate-compact-fluffy-dark}, {aggregate/aggregate-compact-grey\_to\_rhizaria}, {aggregate/aggregate-compact-small-grey}, {compact/compact-light-grey\_to\_globule}, {cnidaria/jellies-w-dots\_to\_badfocus}, {globule/globule\_to\_badfocus}, {globule/globule\_to\_sphere\_thorn}, {globule/small-globule}
  \item[artefact\_badfocus (8):] {artefact/badfocus}, {badfocus/badfocus\_to\_aggregate}, {badfocus/badfocus\_to\_oversegmented}, {bubbles/bubbles-hexagonal\_to\_badfocus}, {cut/cut-jellies-quadratic}, {aggregate/aggregate-fluffy-loose-fiber\_to\_oversegmented}, {aggregate/aggregate-fluffy-very-faint\_to\_badfocus}, {fiber/fiber-fluffy\_to\_feces}
  \item[artefact\_bubble (8):] {bubbles/bubbles-hexagonal}, {bubbles/bubbles-two-halfmoons}, {bubbles/bubbles-two-stars}, {compact/compact-almond-grey\_to\_bubble}, {compact/compact-dark-twins\_to\_bubble-stars}, {compact/compact-grey}, {compact/compact-w-fluffy-surrounds}, {fiber/fiber-bended\_to\_tuft-sharp-ending}
  \item[not-living\_feces (8):] {compact/compact-angled-dark}, {feces/feces-dark-straight\_to\_trichodesmium-tuft}, {feces/feces-little-bended}, {feces/feces-small-grey}, {feces/feces-straight-grey}, {feces/feces\_to\_trichodesmium-tuft}, {tuft/trichodesmium-tuft-dark\_to\_feces}, {unknown/long-even}
  \item[Metazoa\_Annelida (9):] {aggregate/aggregate-fluffy-light-w-flota}, {aggregate/aggregate-long-fluffy-dark\_to\_fiber}, {fiber/fiber-bended-long-slightly-bundled}, {fiber/fiber-bundle-large\_to\_aggregate}, {fiber/fiber-large-long-fluffy}, {metazoa/polychaeta}, {polychaeta/long-bended-worms}, {polychaeta/worms\_to\_badfocus}, {mix/detritus\_to\_chaetognatha}
  \item[detritus\_fiber (10):] {cut/cut-fibers\_to\_turbid}, {aggregate/aggregate-fluffy-sinker\_to\_fiber-bundle}, {feces/feces-bended-lengthy}, {fiber/fiber-bended-fluffy}, {fiber/fiber-bended-thin}, {fiber/fiber-long-slightly-bended-multiple}, {fiber/fiber-medium-bended\_to\_feces}, {fiber/fiber-straight-w-knot}, {fiber/fiber-thin-w-dots}, {unknown/half-moon-w-dot\_to\_badfocus}
  \item[fluffy\_dark (12):] {aggregate/aggregate-angled-grey\_to\_crustacea}, {aggregate/aggregate-dark-fluffy-ball}, {aggregate/aggregate-fluffy-dark-two-spots}, {aggregate/aggregate-fluffy-dark\_to\_feces}, {aggregate/aggregate-fluffy-grey}, {aggregate/aggregate-fluffy-loose-grey}, {aggregate/aggregate-small-fibers-grey\_to\_compact}, {aggregate/jelly-like-remains-and-dark-spots}, {crustacea/copepoda-dark\_to\_amphipoda}, {copepoda/copepoda-compact-dark}, {crustacea/undefined}, {ctenophora/top-view}
  \item[Trichodesmium\_puff (13):] {\bf trichodesmium/puff}, {aggregate/aggregate-small-feathery\_to\_compact}, {compact/compact-angled-dark\_to\_small-crustacea}, {compact/compact-small-round-grey}, {fiber/fiber-bundle-grey}, {fiber/fiber-bundle-small\_to\_puff}, {fiber/fiber\_to\_puff}, {solitary-black/solitary-black-small}, {rhizaria/solitary-black\_to\_puff}, {puff/puff-large}, {puff/puff-large-small}, {puff/puff-medium}, {puff/puff-small}
  \item[Maxillopoda\_Copepoda (19):] {\bf crustacea/copepoda}, {aggregate/aggregate-thorny-grey}, {compact/compact-thorny-dark\_to\_small-crustacea}, {fiber/fiber\_to\_small-crustacean}, {crustacea/amphipoda-like}, {crustacea/copepoda-like}, {crustacea/copepoda-like\_to\_detritus\_compact\_angled\_grey}, {crustacea/copepoda-like\_to\_ostracoda-like}, {crustacea/copepoda-to-detritus}, {copepoda/calanoida}, {calanoida/dorsal-or-ventral}, {calanoida/side-view}, {copepoda/copepoda-small-feathery}, {copepoda/mixed-view}, {copepoda/small-side-view}, {crustacea/drop-like\_to\_aggregate}, {crustacea/small\_crustacea\_to\_feces}, {unknown/ball-w-tentacles}, {unknown/fiber\_w\_growth\_in\_middle}
  \item[Malacostraca\_Eumalacostraca (25):] {\bf crustacea/shrimp}, {badfocus/shrimp-badfocus\_to\_oversegmented}, {aggregate/aggregate-sinker-large-fluffy}, {aggregate/aggregate-very-large-fluffy}, {aggregate/aggregate\_to\_crustacea-decaying}, {detritus/crustacea-parts}, {fiber/fiber-large-bundle}, {fiber/fluffy\_fiber}, {crustacea/amphipoda}, {copepoda/harpacticoida}, {crustacea/copepoda\_to\_badfocus}, {crustacea/crustacea\_to\_amphipoda-like}, {crustacea/drop-like}, {crustacea/drop-like\_to\_badfocus}, {crustacea/drop-like\_to\_shrimp}, {shrimp/shrimp-decaying-or-ill}, {shrimp/shrimp-front-view}, {shrimp/shrimp-like}, {shrimp/shrimp-tails}, {shrimp/shrimp\_to\_oversegmented}, {crustacea/shrimp\_to\_badfocus}, {shrimp\_to\_copepoda/drop\_like}, {crustacea/side-view}, {metazoa/drop-like}, {mix/crustacea\_to\_badfocus}
\end{description}